%% file: main.tex
\documentclass[10pt,twocolumn,letterpaper]{article}

\usepackage[pagenumbers]{cvpr} 

\usepackage{amsmath}
\usepackage{amssymb}
\usepackage{multirow} 
\usepackage{float}
\usepackage{algorithm}
\usepackage{algorithmic}
\usepackage{tabularx}
\usepackage{arydshln}
\usepackage{pifont}
\usepackage{caption}
\usepackage{xcolor}
\usepackage{hyperref}
\hypersetup{
    colorlinks=true,   
    urlcolor=cyan,
}

\definecolor{cvprblue}{rgb}{0.21,0.49,0.74}

\newcommand{\chany}[1]{\textcolor{red}{{[C: #1]}}}

\def\paperID{2166} 
\def\confName{CVPR}
\def\confYear{2025}

\title{Data-free Universal Adversarial Perturbation with Pseudo-semantic Prior}

\def\authorBlock{
    \enspace \enspace
    $\text{Chanhui Lee}$ \qquad \qquad \qquad \enspace
    $\text{Yeonghwan Song}$ \qquad \qquad \quad \qquad
    $\text{Jeany Son}$ \\
    $\text{AI Graduate School, GIST}$ \\
    {\tt\small \{as584868, yeonghwan.song\}@gm.gist.ac.kr} \qquad
    {\tt\small jeany@gist.ac.kr}
}
\author{\authorBlock}
\begin{document}
\maketitle
\input{sec/0_abstract_camera}   
\input{sec/1_intro_camera}

\input{sec/2_related}


\input{sec/3_method_camera}

\input{sec/4_Experiments_camera}
\input{sec/5_Conclusion}
{
    \small
    \bibliographystyle{ieeenat_fullname}
    \bibliography{main}
}
\input{sec/X_suppl}


\end{document}

%% file: sec/0_abstract_camera.tex
\begin{abstract}
Data-free Universal Adversarial Perturbation (UAP) is an image-agnostic adversarial attack that deceives deep neural networks using a single perturbation generated solely from random noise without relying on data priors.
However, traditional data-free UAP methods often suffer from limited transferability due to the absence of semantic content in random noise.
To address this issue, we propose a novel data-free universal attack method that recursively extracts pseudo-semantic priors directly from the UAPs during training to enrich the semantic content within the data-free UAP framework.
Our approach effectively leverages latent semantic information within UAPs via region sampling, enabling successful input transformations—typically ineffective in traditional data-free UAP methods due to the lack of semantic cues—and significantly enhancing black-box transferability.
Furthermore, we introduce a sample reweighting technique to mitigate potential imbalances from random sampling and transformations, emphasizing hard examples less affected by the UAPs.
Comprehensive experiments on ImageNet show that our method achieves state-of-the-art performance in average fooling rate by a substantial margin, notably improves attack transferability across various CNN architectures compared to existing data-free UAP methods, and even surpasses data-dependent UAP methods.
Code is available at:~\url{https://github.com/ChnanChan/PSP-UAP}.

\end{abstract}

%% file: sec/1_intro_camera.tex
\section{Introduction}
\label{sec:intro}


Deep neural networks (DNNs) have become widely used in computer vision, achieving remarkable performance across a diverse range of tasks, such as image classification~\cite{res152, inc_v3}, object detection~\cite{yolo, faster}, semantic segmentation~\cite{u-net}, and visual tracking~\cite{siam, fairmot}.
Despite these successes, DNNs are vulnerable to carefully crafted, imperceptible perturbations in input data, causing the model to make highly confident yet incorrect predictions.
This vulnerability poses significant challenges for deploying DNNs in critical applications, such as autonomous driving~\cite{autonomous} and security systems~\cite{security}, and has led to increased research into adversarial attacks that generate adversarial examples.


\begin{figure}[!t]
\centering    \includegraphics[width=\linewidth]{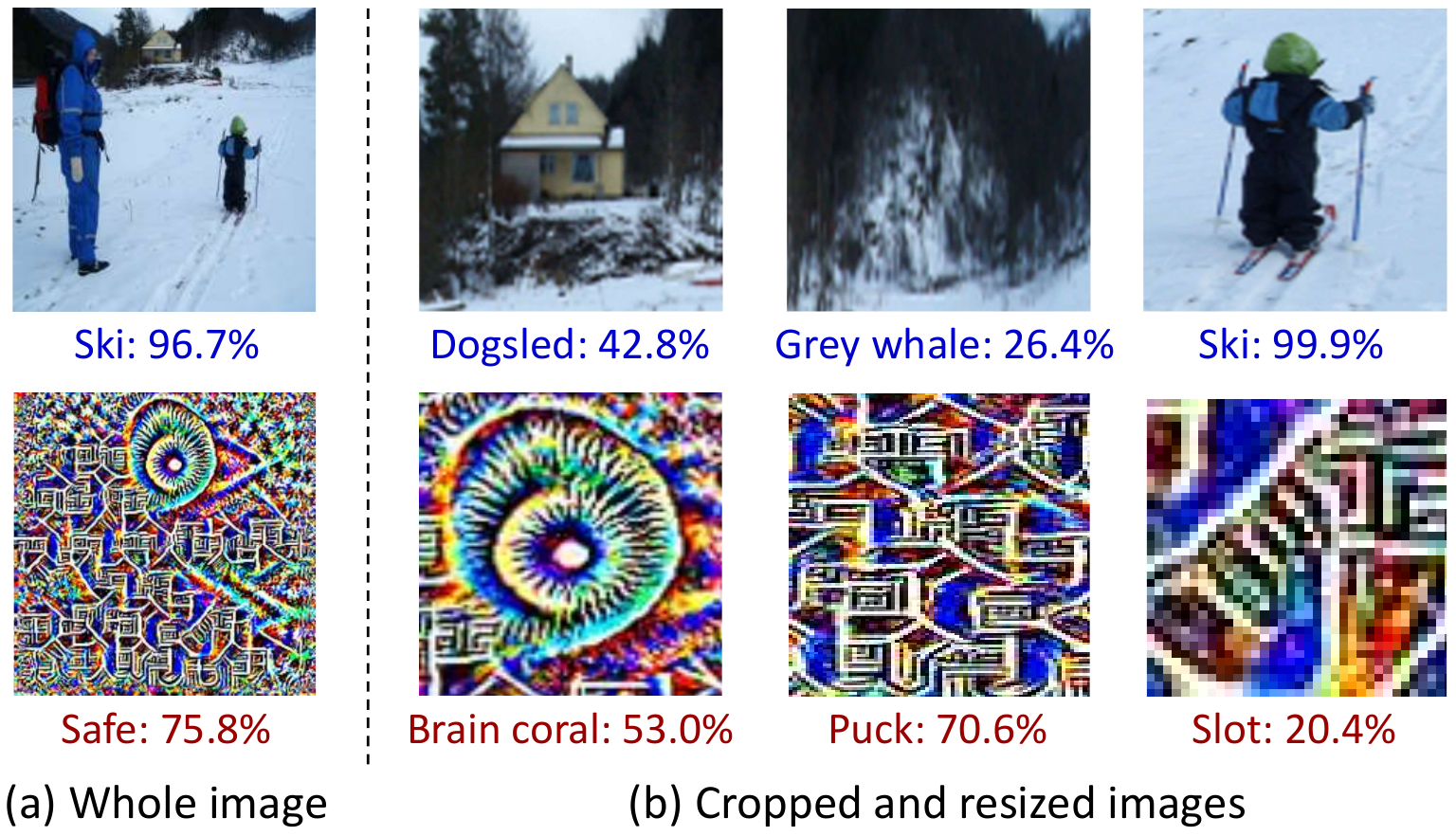}
    \caption{ Diverse semantic contents in both a real-image and UAP: 
    {(a)} Whole images from the ImageNet dataset (top) and our generated data-free UAP (bottom) using DenseNet-121, shown at iteration 900 during the training phase. The Top-1 class and its score are shown below each image. 
    {(b)} Cropped regions from the whole image (top) and our UAP (bottom). Those regions contain diverse semantics that differ from the class of the original images.
    }
    \label{fig:fig_1}
\end{figure}

To craft the adversarial examples with high transferability across various DNN architectures, there have been many adversarial attack methods using a specific target image~\cite{FGSM, Deepfool, mifgsm, l-bfgs, generating, i-fgsm}.
However, these methods generate a unique perturbation for each target image, which is time-consuming, impractical for real-world scenarios, and limits their generalization to other images.
To tackle this limitation, the Universal Adversarial Perturbation (UAP)~\cite{UAP} introduced an image-agnostic attack that generates a single image-agnostic adversarial perturbation, which is capable of attacking a wide range of unknown images.
Many studies~\cite{sga-uap, spgd, nag, gap} focus on developing data-dependent UAPs that target unknown models, thereby enhancing transferability across diverse and unseen scenarios.
While these UAPs deceive diverse categories of images with a single perturbation, they still rely on large-scale data samples and their labels from the target domain to capture diverse semantics, such as the ImageNet~\cite{Imagenet} dataset.

Accessing data priors from the target domain is often impractical, leading to recent interest in data-free UAP methods~\cite{FFF,gd-uap,TRM,pd-ua,AAA,at-uap,cosine}.
Data-free UAP poses a greater challenge than conventional data-dependent UAP generation tasks, as it restricts the employment of any prior knowledge of the target domain dataset.
Prior works~\cite{FFF, gd-uap, TRM} have attempted to craft UAPs from random noise without any dataset, by maximizing activations in convolutional neural networks (CNNs) layers. 
However, these methods solely rely on random priors, such as Gaussian noise or jigsaw patterns, which lack semantic information and thus offer limited transferability to unseen models.
To overcome this limitation, several works~\cite{AAA, at-uap} utilize auxiliary data samples generated by optimizing against the outputs of a surrogate model.
Although these methods allow the use of semantic information in synthetic data, the crafted UAPs often show inferior transferability due to overfitted data to the surrogate model.

In this paper, we explore how to leverage semantic information directly from the UAP itself, without any dataset priors, to address these challenges.
Our approach is inspired by the observation that even a single generated UAP contains diverse semantic information as well as its dominant label, as shown in Figure~\ref{fig:fig_1}.
We find that the generated UAP encodes diverse semantic features, similar to a real-world image with various semantic contents across different regions.
For example, in Figure~\ref{fig:fig_1}, while the whole UAP is predicted as `Safe' due to its tendency to have a dominant label, cropped regions within the UAP are predicted as classes like `Brain coral,' `Puck,' and `Slot.' 
This observation motivates us to utilize UAP as a semantic prior for training within a data-free UAP framework.

Inspired by this insight, we propose PSP-UAP, a novel data-free UAP method that generates pseudo-semantic priors from a UAP during training.
To capture more diverse semantics in pseudo-semantic priors, we randomly crop and resize regions to extract semantic samples and treat them as images to be fooled. 
This approach effectively addresses the data-free constraint in UAP generation by leveraging richer semantic information inherent in the UAP, rather than relying solely on random noise.
To improve attack transferability, we further incorporate input transformations~\cite{xie2019improving, L2T}, commonly used in image-specific adversarial attacks, into our data-free UAP framework.
This strategy has not been explored in existing data-free UAP methods, as random priors lack semantic information, limiting its effectiveness.
In contrast, our pseudo-semantic prior contains richer semantic content, thereby enabling improved transferability to unknown models through input transformations.
Moreover, since semantic samples obtained through random cropping and transformation vary in informativeness, we introduce a sample reweighting that prioritizes hard examples, which are less effectively deceived by the current UAP, to improve the overall effectiveness of the generated UAPs.


The main contributions can be summarized as follows:
\begin{itemize}
    \item We propose PSP-UAP, a novel data-free universal attack method that generates {{pseudo-semantic priors}} from the UAP itself, using inherent semantic information of the UAP as an alternative data source during training.
    \item We are the first to incorporate input transformations into the data-free UAP framework by leveraging pseudo-semantic priors with diverse semantic cues, boosting transferability across various CNN architectures in black-box settings.
    \item 
    Our sample reweighting prioritizes challenging examples during UAP training, by reducing the influence of uninformative samples produced by random sampling in our pseudo-semantic prior and input transformations.
    \item Our method achieves outstanding performance over the state-of-the-art data-free UAP methods by a substantial margin, and even outperforms existing data-dependent UAP methods.
\end{itemize}

%% file: sec/2_related.tex
\section{Related Work}
\label{sec: Related work}

\paragraph{Data-dependent Universal Attack.} Data-dependent UAPs aim to generate a single perturbation that misleads any image sample.
UAP~\cite{UAP} first proposed finding the minimal universal adversarial perturbation at each step by DeepFool~\cite{Deepfool} method. 
SPGD-UAP~\cite{spgd} combined the stochastic gradient method with the projected gradient descent (PGD)~\cite{pgd} attack method.
SGA-UAP~\cite{sga-uap} used stochastic gradient aggregation in mini-batch to address gradient vanishing and quantization errors.
AT-UAP~\cite{at-uap} integrated image-specific and image-agnostic attacks to improve the robustness of universal perturbation.
NAG~\cite{nag} and GAP~\cite{gap} applied generative adversarial frameworks to craft perturbations.
Although these works~\cite{spgd, sga-uap, nag, gap} increase the transferability of black-box attacks, they require the training dataset, making it impractical when the adversary does not have any prior of the target domain.


\begin{figure*}[!t]
\centering
    \includegraphics[width=\linewidth]{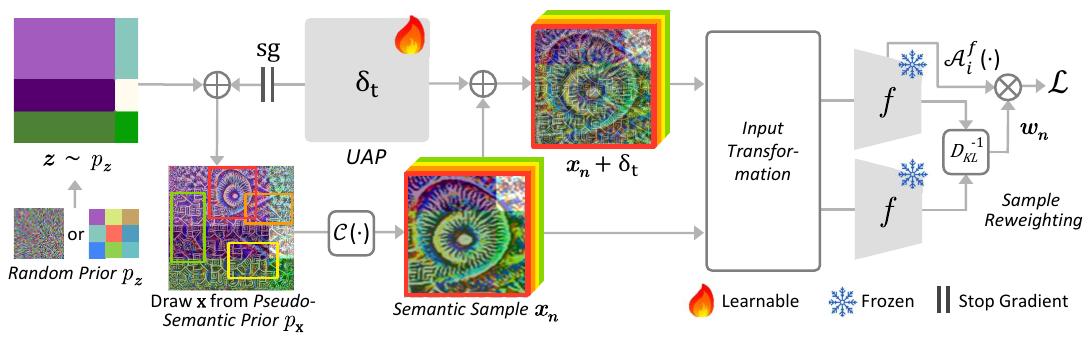}
    \vspace{-0.5cm}
    \caption{Overall pipeline of the proposed PSP-UAP. The pseudo-semantic prior is created by adding random noise to the UAP. Semantic samples are then generated by randomly cropping and resizing the pseudo-semantic prior. Input transformation is applied to both adversarial and clean versions of the semantic samples to calculate sample reweighting. Finally, the loss is defined as the product of sample reweighting and the activations of the semantic samples, from which gradients are computed to update the model.}
    \label{fig:framework}
    
\end{figure*}

\vspace{-0.2cm}
\paragraph{Data-free Universal Attack.} 
Data-free universal attack aims to craft UAPs without any dataset to be used for training, thereby alleviating the data access requirement presented in data-dependent universal attacks.
Generating UAP without prior knowledge of the target domain is more practical and suitable for real-world applications.
Fast Feature Fool (FFF)~\cite{FFF} first proposed a data-free universal adversarial attack by maximizing the feature activation values of all CNN layers.
Generalizable Data-free UAP (GD-UAP)~\cite{gd-uap} improved FFF~\cite{FFF} attack method with saturation check strategy on the training stage. 
AT-UAP~\cite{at-uap} also performed experiments in a data-free manner by applying adversarial attacks to random noise.
Prior-Driven Uncertainty Approximation (PD-UA)~\cite{pd-ua} introduced an attack method to train UAP by maximizing the uncertainty approximation of the model with the prior patterns, and Cosine-UAP~\cite{cosine} proposed the minimizing cosine similarity to craft UAPs in a self-supervised manner.
AAA~\cite{AAA} crafted class impressions with logits to train a generative model to optimize UAPs. 
TRM-UAP~\cite{TRM} increased the ratio of positive and negative activations on the shallow convolution layers and adapted curriculum learning to enhance attack transferability stably.
Despite employing various methods to generate UAPs in a data-free setting, they face significant challenges due to the lack of information on both the target models and domains.
Additionally, these works substantially rely on random priors to generate UAPs, and auxiliary data directly utilizes label information, leading to overfitting on surrogate models.

\vspace{-0.3cm}
\paragraph{Input Transformation Attack.}
Input transformation methods have emerged as one of the effective ways to improve attack transferability in image-specific adversarial attacks.
Diverse input method (DIM)~\cite{xie2019improving}, translate invariant method (TIM)~\cite{dong2019evading}, and scale invariant method (SIM)~\cite{lin2019nesterov} revealed the DNN models' invariant properties to transformations such as resizing, translation, and scaling before the gradient calculation.
SIA~\cite{SIA} applied various transformations to the input image while maintaining its overall structure.
\textit{Admix}~\cite{admix} created admixed images by blending a small fraction of images from different categories into the input image.
Block shuffle and rotation (BSR)~\cite{bsr} randomly shuffled and rotated the sub-blocks of the input image to reduce the variance in attention heatmaps across different models.
L2T~\cite{L2T} used reinforcement learning to increase the diversity of transformed images by selecting the optimal transformation combinations.
To fully leverage our pseudo-semantic prior, we incorporate input transformations into our data-free UAP method to further enhance black-box transferability.


%% file: sec/3_method_camera.tex
\section{Methodology}

In this section, we present our motivation and approach for generating the pseudo-semantic prior.
We then describe the input transformation applied to semantic samples derived from the pseudo-semantic prior, followed by our sample reweighting strategy for optimizing UAP.

\subsection{Preliminaries of data-free UAP}

Universal adversarial attacks aim to optimize a single perturbation $\delta$ using a model $f$ that effectively deceives most of the samples $I$ in the target domain dataset, with the pixel intensities of $\delta$ restricted by a constraint parameter $\epsilon$:
\begin{align}
    f (I + \delta) \neq f (I), \text{~~~s.t.~~} \| \delta \|_{\infty} \leq \epsilon.
    \label{eq:obj}
\end{align}
However, in data-free settings where the target dataset is inaccessible, UAPs are typically trained using simple random priors, such as Gaussian noises or jigsaw images~\cite{cosine, TRM}.
Given these random priors~$p_{z}$, GD-UAP~\cite{gd-uap} introduced an activation maximizing loss as follows:

\begin{gather}
    \mathcal{L} = -\mathbb{E}_{{z}\sim p_{{z}}}\sum_{i=1}^L{\log{\|\mathcal{A}^f_{i}({z}+ \delta)}\|_2},\label{eq:GDUAP}\\
    \text{s.t.~~} \| \delta \|_{\infty} \leq \epsilon \nonumber, 
\end{gather}
where $\mathcal{A}^f_{i}(\cdot)$ indicates the activation of the $i$-th layer of the surrogate network $f$, $L$ denotes the number of layers in $f$, and ${z}$ represents pseudo-data sampled from a simple random prior distribution, $p_{{z}}$.
This loss is designed to overactivate features extracted from multiple convolutional layers of the surrogate model without input images.
Consequently, the distorted activation interferes with feature extraction, leading CNN models to make incorrect predictions~\cite{FFF, gd-uap}.

\subsection{Pseudo-Semantic Prior}

Although data-free UAP methods~\cite{gd-uap, TRM} use random priors, such as Gaussian noises or artificial jigsaw puzzles, to mimic the statistical properties of image datasets, they are still limited by a lack of semantic information.
Furthermore, since UAPs are trained by maximizing activations in network layers, they tend to overfit to surrogate models.  
Optimizing UAPs without semantic content and relying on  activation or outputs of surrogate networks, reduces their effectiveness in disrupting real images in unseen models, leading to degraded performances in black box transferability.

To address these issues, we aim to enhance the semantic content within a data-free UAP framework, inspired by previous works~\cite{UAP, sga-uap, cosine, TRM} that demonstrate the presence of dominant labels in generated UAPs.
For instance, early works~\cite{UAP, sga-uap} observed that untargeted UAPs often cause misclassification toward a dominant label, a property that also holds in data-free settings~\cite{TRM}.
Cosine-UAP~\cite{cosine} further showed that the logit distribution of UAPs tends to dominate that of the input data $x$.
This suggests that, although the UAP is a subtle perturbation, it behaves like a single image with strong semantic information, guiding the model toward classification with a dominant label.

\begin{figure}[t]
\centering
    \includegraphics[width=\linewidth]{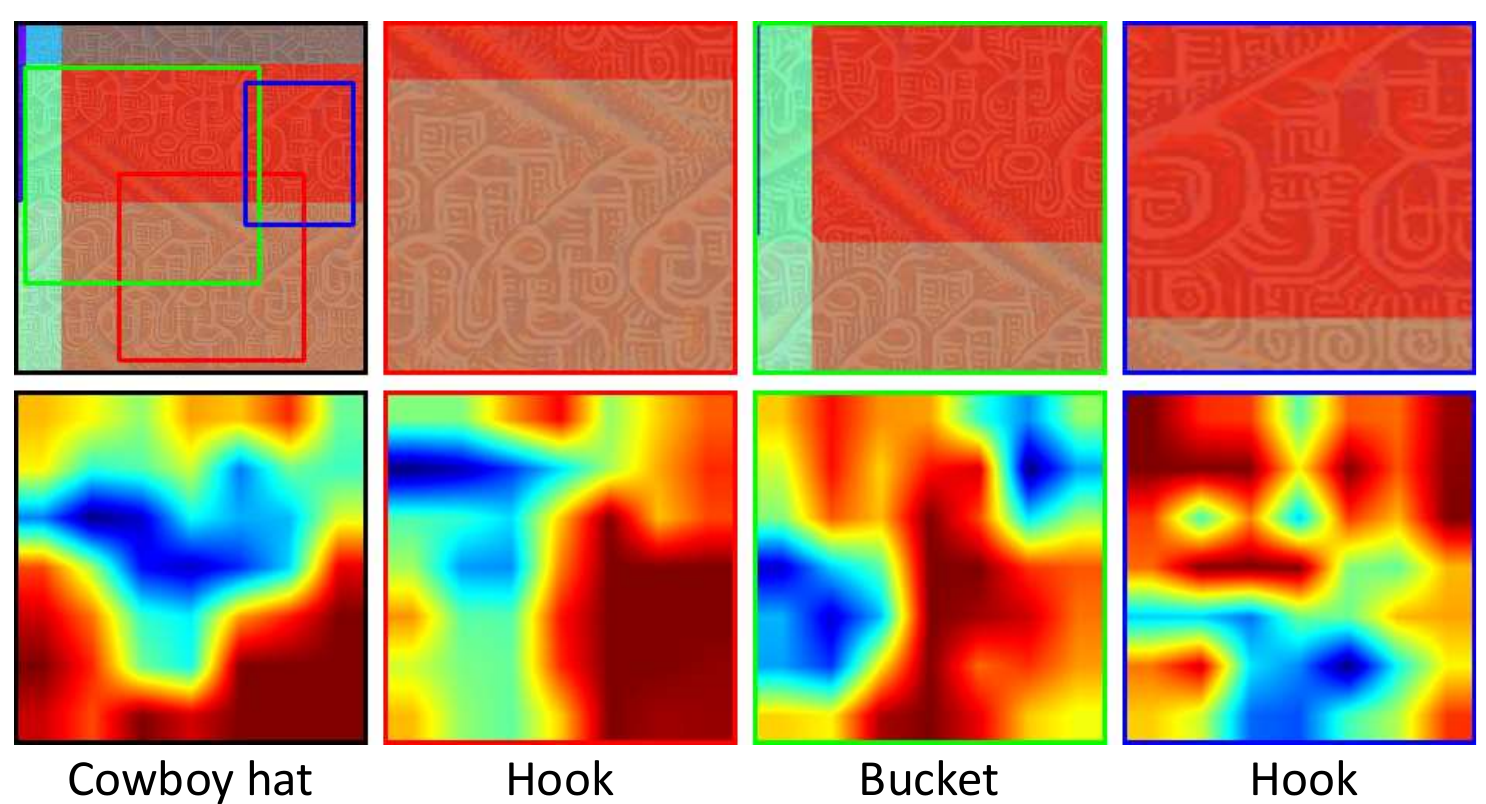}
    \caption{
    Semantic samples derived from an adversarial example during training, along with their predicted labels and GradCAM heatmaps from Dense-121. 
    Despite originating from the same example, the variations in predicted labels and heatmaps indicate that these semantic samples capture diverse semantic features.
    }
    \label{fig:fig_3}
\end{figure}

Inspired by this observation, we leverage the inherent semantic information in UAPs by treating the combination of UAP and random noise as a single image, termed the \textbf{\textit{pseudo-semantic prior}}, to resolve the lack of semantic content in data-free UAP training.
As shown in Figure~\ref{fig:fig_1} and Figure~\ref{fig:fig_3}, the generated UAPs exhibit diverse semantic labels across different regions.
Although {regions within the UAP are} classified under the same label, the attention heatmaps generated by Grad-CAM~\cite{gradcam} show distinct patterns, suggesting that diverse semantic patterns are embedded within the UAP.

Building on the above insight, we generate pseudo-data samples from the pseudo-semantic prior to enrich the semantic content, which we refer to as \textbf{\textit{semantic samples}} $x_n$:
\begin{gather}
    \mathbf{x} \sim p_{\mathbf{x}|p_{{z}}, \delta_t} = \{{z} + \delta_t \vert  {z} \in p_{{z}}\}, \label{SP} \\
    \{{x}_1, {x}_2, ..., {x}_N\} = \mathcal{C}(\mathbf{x}; N),
    \label{crop_resize}
\end{gather}
where ${z}$ and $\delta_t$ denote random noise sampled from $p_{{z}}$ and the UAP being trained in $t$-th iteration, respectively.
The pseudo-semantic prior $p_{\mathbf{x}}$ denotes a set of adversarial examples with $\delta_t$ derived from the random prior distribution.
$\mathcal{C}$ is a sampler that draws $N$ numbers of semantic samples ${x}_n$ from $p_{\mathbf{x}}$ by applying crop and resize operations.
Specifically, we randomly crop a region of $\mathbf{x}$, the sum of random noise $z$ and the UAP $\delta_t$, and resize it to the original scale of the UAP size.
We believe that leveraging the generated semantic information embedded in different regions of the UAP provides more effective guidance for successfully attacking target features than relying solely on random noise.

\subsection{Input Transformation}
\label{subsec:input_transformation}
To enhance black-box attack transferability, we incorporate input transformation techniques into our semantic samples. 
While input transformation is commonly used in image-specific adversarial attacks, it has been less explored in data-free UAPs due to the limited semantic information in random priors. 
In our method, however, the semantic samples generated from pseudo-semantic priors contain diverse semantic cues, making input transformations more effective in a data-free setting.

Following L2T~\cite{L2T}, which shows that rotation, scaling, and shuffling are particularly effective for improving attack transferability, we randomly select one of these transformations and apply it to each semantic sample.
For rotation, the angle $\alpha$ is drawn from a truncated normal distribution within the range $-\theta \leq \alpha \leq \theta$.
Scaling is applied with a uniform distribution within the bounds $\beta_{low} \leq \beta \leq \beta_{high}$.
Shuffling involves randomly rearranging $m\times m$ blocks.
During optimization, applying input transformations to our PSP-UAP increases the variation of semantic samples and enhances the black-box attack transferability.

\subsection{Sample Reweighting}
Random processes involved in drawing semantic samples from the pseudo-semantic prior and applying input transformations lead to an imbalance in difficulty due to variations in semantic content; some samples are easily fooled by the UAP, while others are more difficult to deceive.
To tackle this, we propose a new sample reweighting method that prioritizes harder-to-fool samples. 

Specifically, we compute a weight for each sample using the KL-divergence between the transformed input and its adversarial counterpart during training.
We define the original distribution~$P$ and the adversarial distribution~$Q$ for each semantic sample and its corresponding adversarial example generated by the current UAP~$\delta$ as follows:
\begin{align}
    &P({x}_n) = f(T({x}_n)) \label{eq:ppp},\\
    &Q({x}_n) = f(T({x}_n+\delta_t)) \label{eq:qqq},
\end{align}
where $f(\cdot)$ denotes the temperature-scaled softmax output of a surrogate model {and {and $T(\cdot)$} represents a randomly selected input transformation} as described in Sec.~\ref{subsec:input_transformation}.
We then compute the weights for each semantic sample using the KL-divergence:
\begin{align}
    &w_n = D_{KL}(P({x}_n)\|Q({x}_n))^{-1}.
    \label{eq:weight_1}
\end{align}
The large KL-divergence value indicates that $\delta_t$ has significantly altered the distribution of ${x}_n$, whereas a small value suggests an ineffective attack.
Thus, we take the reciprocal of KL-divergence values to assign greater weights to semantic samples where the attack results in minimal distributional change.
We reweight the semantic samples using the weights generated from Eq.~\eqref{eq:weight_1}, considering the influence of the UAP on each sampled ${x}_n$, which enables us to optimize the UAP effectively.



\input{table/algorithm}

\subsection{Overall Loss}
We integrate the pseudo-semantic prior, input transformation, and sample reweighting into Eq.~\eqref{eq:GDUAP} to optimize the proposed PSP-UAP.
The final loss function of our PSP-UAP is defined as follows:
\begin{gather}
   \mathcal{L} = -\mathbb{E}_{\mathbf{x} \sim p_{\mathbf{x}}} \sum_{n=1}^N \sum_{i=1}^{l}{\log{(w_n \|\mathcal{A}^f_{i}(T({x}_n+ \delta_t))}\|_2)},
    \label{eq:overall loss}
\end{gather}
where $\mathcal{A}^f_{i}(\cdot)$ denotes the activation of the $i$-th layer in network $f$, ${x}_n$ represent the $n$-th semantic sample extracted from the pseudo-semantic prior $p_{\mathbf{x}}$, $\delta_t$ denotes the UAP at the $t$-th iteration,
$w_n$ is the weight of ${x}_n$ from Eq.~\eqref{eq:weight_1}, $l$ indicates the number of the convolutional layers used in the activation sum,
$T(\cdot)$ denotes a randomly selected transform for input transformation, and $N$ is the number of semantic samples extracted from one pseudo-semantic prior.

By optimizing the overall loss, we fully leverage the generated semantic samples as data to produce a highly transferable UAP, even without prior knowledge of the target domain. 
The overall PSP-UAP framework and detailed algorithm are shown in Figure~\ref{fig:framework} and Algorithm~\ref{alg:psp-uap}, respectively.

\input{table/white_box}

%% file: table/algorithm.tex
\begin{algorithm}[t]
\caption{Pseudo-semantic Prior Universal Attack}
\label{alg:psp-uap}
\textbf{Input:} Surrogate model $f$, number of semantic samples $N$, maximum perturbation magnitude $\epsilon$, learning rate $\eta$, maximum iteration number $T$, convergence threshold $F_{\text{max}}$, validation test hyperparameter $H$, saturation threshold $r$. \\
\textbf{Output:} Universal adversarial perturbation $\delta$.

\begin{algorithmic}[1]
    \STATE Initialize $\delta_0 \sim \mathcal{U}(-\epsilon, \epsilon), t=0, F=0$
    \WHILE{$t < T$ and $F < F_{max}$}
        \STATE $t = t + 1$
        \STATE Generate the random noise set ${z} \sim p_{{z}}$ 
        \STATE Update pseudo-semantic prior $p_{\mathbf{x}}$ with $\delta_t$ via Eq.~\eqref{SP}
        \STATE Sample $N$ semantic samples ${x}_n$ via Eq.~\eqref{crop_resize}
        \STATE Select and apply transformation $T \in$ \{\textit{rotation}, \textit{scaling}, \textit{shuffling}\}
        \STATE Compute the weight ${w}$ via Eq.~\eqref{eq:ppp},~\eqref{eq:qqq}, and \eqref{eq:weight_1}
        \STATE Calculate the gradient $\nabla \mathcal{L}$ of the loss in Eq.~\eqref{eq:overall loss}
        \STATE Update $\delta_t = \delta_{t-1} + \eta \cdot \nabla \mathcal{L}$
        \STATE  Clip $\delta_t = \text{min}(\epsilon, \text{max}(\delta_t, -\epsilon))$
        \STATE Compute the saturation rate $\hat{r}$ and adjust $\delta_t$ if $r < \hat{r}$
                \IF{$t \% H == 0$}
        \STATE Conduct the fooling rate test $FR$
        \IF{$FR$ is not the best fooling rate}
            \STATE $F = F + 1$
        \ENDIF
    \ENDIF
    \ENDWHILE
    \RETURN $\delta_t$
\end{algorithmic}
\end{algorithm}

%% file: table/white_box.tex
\begin{table}[t]
  \centering
  \footnotesize 
  \setlength{\tabcolsep}{2.5pt} 
  \renewcommand{\arraystretch}{1.1} 
  \begin{tabular}{lccccc c}
    \toprule
    Attack & AlexNet & VGG16 & VGG19 & RN152 & Google & ~Avg.~ \\
    \midrule
    FFF~\cite{FFF} & 80.92 & 47.10 & 43.62 & - & 56.44 & - \\
    AAA~\cite{AAA} & 89.04 & 71.59 & 72.84 & 60.72 & 75.28 & 73.89 \\
    GD-UAP~\cite{gd-uap} & 85.24 & 90.01 & 87.34 & 45.96 & 45.87 & 70.88 \\
    PD-UAP~\cite{pd-ua} & - & 70.69 & 64.98 & 46.39 & 67.12 & - \\
    Cosine-UAP~\cite{cosine} & 91.07 & 89.48 & 86.81 & 65.35 & \textbf{87.57} & 84.05 \\
    AT-UAP-U~\cite{at-uap} & \textbf{96.66} & 94.50 & 92.85 & 73.15 & 82.60 & 87.95 \\
    TRM-UAP~\cite{TRM} & 93.53 & 94.30 & 91.35 & 67.46 & 85.32 & 86.39 \\
    \textbf{PSP-UAP (Ours)} & 91.77 & \textbf{96.26} & \textbf{94.65} & \textbf{85.65} & 81.43 & \textbf{89.95} \\
    \bottomrule
  \end{tabular}
  \caption{$FR$ (\%) of our PSP-UAP and other data-free universal attack methods for white-box attacks.}
  \vspace{-0.3cm}
  \label{tab:white_box}
\end{table}

%% file: sec/4_Experiments_camera.tex
\input{table/black_box_alex}
\input{table/black_box}

\input{table/data-dependent}

\section{Experiments}
\paragraph{Experimental Setup.} 
We follow the experiment setup in existing data-free universal attacks~\cite{gd-uap, TRM} to evaluate the performance of our PSP-UAP.
We evaluate the proposed method on ImageNet~\cite{Imagenet} validation set with five ImageNet pre-trained CNN models, AlexNet~\cite{alex}, VGG16~\cite{vgg}, VGG19~\cite{vgg}, ResNet152 (RN152)~\cite{res152}, and GoogleNet~\cite{google}, which are commonly used in data-free UAP methods.
We further explore four additional CNN models including DenseNet121~\cite{dense}, MobileNet-v3-Large~\cite{mobile}, ResNet50~\cite{res152}, and Inception-v3~\cite{inception}, pre-trained on ImageNet dataset.


\vspace{-0.2cm}
\paragraph{Evaluation Metrics.} To effectively evaluate the attack performance of our proposed method, we use the fooling rate $(FR)$ which is widely used in universal attacks ~\cite{UAP, TRM}.
$FR$ indicates the proportion of samples with label changes when applying UAP.

\vspace{-0.2cm}
\paragraph{Baselines.} The proposed method is compared with the following data-free universal attacks, including FFF~\cite{FFF}, GD-UAP~\cite{gd-uap}, PD-UA~\cite{pd-ua}, Cosine-UAP~\cite{cosine}, AT-UAP~\cite{at-uap}, and TRM-UAP~\cite{TRM}.
Since AT-UAP includes both data-free (AT-UAP-U) and data-dependent (AT-UAP-S) versions, we evaluate both in our experiments.
We also compared our method with SGA-UAP~\cite{sga-uap} which is one of the state-of-the-art data-dependent universal attacks.

\vspace{-0.2cm}
\paragraph{Implementation Details.} Our experiments are implemented on PyTorch with a single NVIDIA A6000 GPU.
We set $\epsilon$ $=$ $10/255$ to restrict $\ell_{\infty}$-norm, the maximum iteration $T$ as $10,000$, and the saturation threshold $r$ to $0.001\%$, following the setting in TRM-UAP~\cite{gd-uap, TRM}.
For input transformations, we set $\theta$ $=$ $6$ for rotation, $\beta_{low}$ $=$ $0.8$ and $\beta_{high}$ $=$ $4$ for scaling, and $m$ $=$ $2$ for random shuffling.
Moreover, the number of semantic samples $N$ is set to $10$.
We define the temperature parameter $\tau \in \{1.0, 5.0, 5.0, 3.0, 5.0\}$ corresponding to AlexNet, VGG16, VGG19, ResNet152, GoogleNet.
For the additional CNN models, we set $\tau \in \{3.0, 10.0, 2.0, 3.0\}$ corresponding to ResNet50, DenseNet121, MobileNet-v3-Large, Inception-v3.
We follow the same curriculum learning and saturation check strategy of TRM-UAP.
We set different ratios to use the activations of intermediate layers across different models.
To ensure a fair evaluation, we find the optimal parameters for TRM-UAP on RN50, DN121, MB-v3-L, and Inc-v3 through our best efforts.


\subsection{Evaluation on White-Box Attack}
We first evaluate UAPs generated by our PSP-UAP on various CNN models under the white-box setting.
We compare the attack performance of our UAPs with other data-free universal attacks on the ImageNet validation set, as shown in Table~\ref{tab:white_box}.
While the $FR$ on AlexNet and GoogleNet is slightly lower than other methods, our approach achieves the highest average $FR$ across all universal attack methods.
Notably, the improvement on ResNet152 is substantial, with a $12.5\%$ increase in the $FR$, 
demonstrating that our PSP-UAP performs exceptionally well, particularly on deeper and more complex CNN models.


\subsection{Evaluation on Black-Box Attack}

\paragraph{Comparison with SoTA Data-free UAPs.}
We evaluate the transferability of our UAPs on commonly used CNN models in the black-box scenario.
Table~\ref{tab:black_box_combined} represents the attack performance across different settings, with columns representing target models and rows indicating surrogate models to craft the UAPs. 
As shown in Table~\ref{tab:black_box_combined}, PSP-UAP achieves superior results than other data-free attack methods across most models, with performance comparable to, but slightly below, TRM-UAP on GoogleNet. 
Notably, in strictly black-box settings (excluding the white-box scenario where GoogleNet attacks itself), PSP-UAP achieves an average $FR$ of 70.1\% compared to TRM-UAP’s 69.6\%, demonstrating better transferability.

\vspace{-0.2cm}
\paragraph{Extended Evaluation with Additional CNN Models.}
To further explore the effectiveness of the proposed PSP-UAP, we conduct additional experiments on widely used CNN models, including ResNet50, DenseNet121, MobileNet-v3-Large, and Inception-v3.
We compare the attack performance of PSP-UAP with TRM-UAP as shown in Table~\ref{tab:black_box_advance}.
Note that we limit the comparison to TRM-UAP since the public code for AT-UAP has not yet been released.
The results demonstrate that our PSP-UAP consistently outperforms TRM-UAP in terms of $FR$ with a substantial margin.
The substantial improvement demonstrates PSP-UAP's strong generalization across diverse CNN models, highlighting the robustness of our approach.

\vspace{-0.2cm}
\paragraph{Comparison with Data-dependent UAPs.}
To verify whether our method effectively alleviates the lack of prior knowledge, we compare our method to state-of-the-art data-dependent universal attacks in the black-box scenario.
As shown in Table~\ref{tab:black_box_data_dep}, the $FR$ of white-box attacks is inevitably higher for SGA-UAP and AT-UAP-S, as they fully utilize the target domain dataset.
On the other hand, PSP-UAP exhibits superior transferability, as observed in the black-box setting.
Our method outperforms by achieving a higher average $FR$ across most models, surpassing data-dependent approaches.
Furthermore, even in cases where some $FR$ results are lower, they do not fall significantly behind the data-dependent universal methods.
These results indicate that our semantic samples prove to be effective as data for training the UAP, successfully overcoming the limitation posed by the absence of a target domain.

\subsection{Ablation Study} We conduct ablation experiments to explore the effectiveness of the proposed pseudo-semantic prior, sample reweighting, and applying input transformation techniques on the semantic samples.
In the following experiments, we generate UAPs on ResNet152 and evaluate them across various CNN models (\eg, AlexNet, VGG16, VGG19, ResNet152, and GoogleNet).

\begin{figure*}[t]
\centering
\begin{minipage}[t]{0.48\linewidth}
    \includegraphics[width=\linewidth]{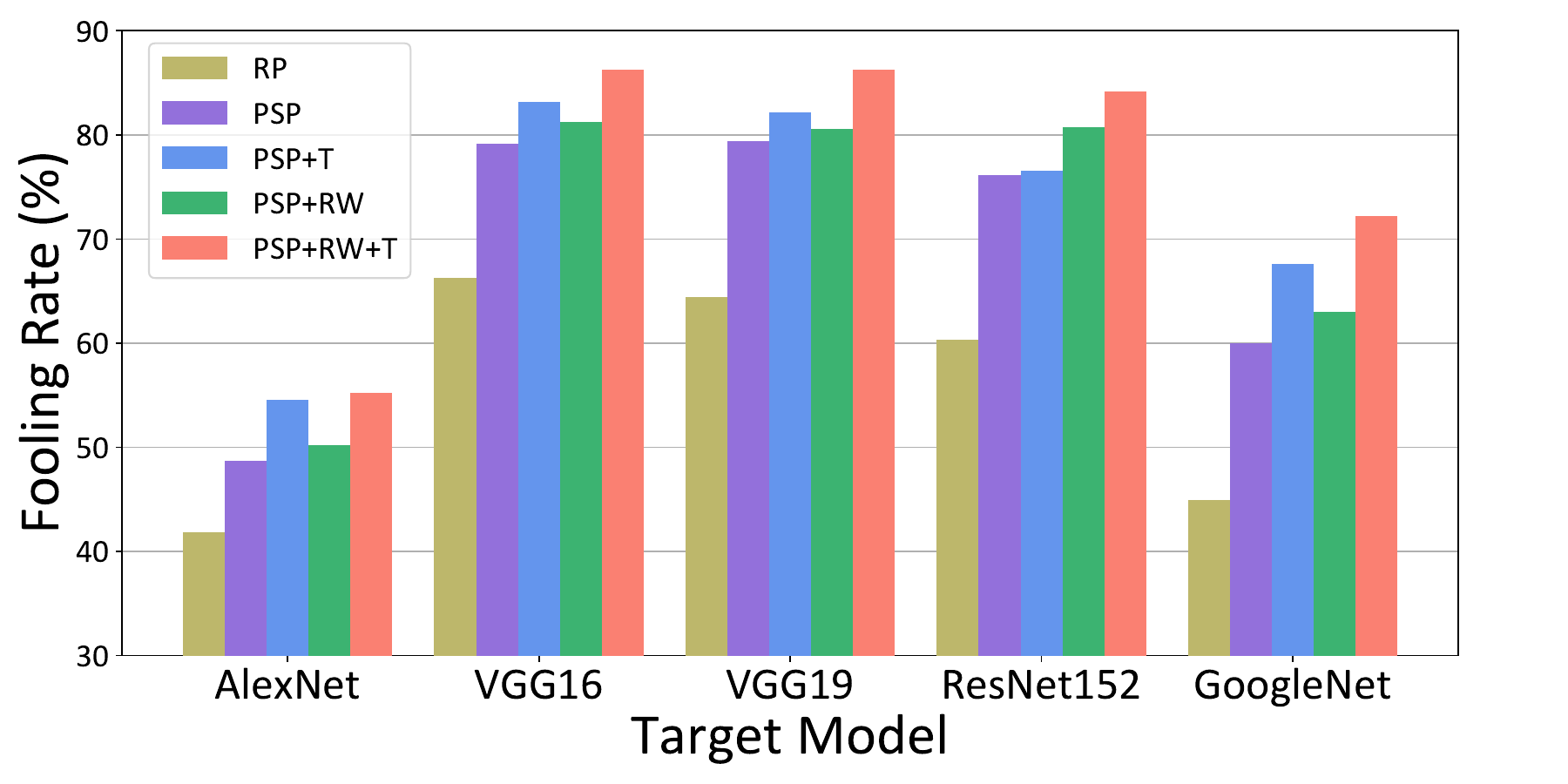}
    \vspace{-0.8cm}
    \caption{
    Ablation study on each proposed component in PSP-UAP.
    RP and PSP refer to training a UAP using random noises and semantic samples drawn from pseudo-semantic prior, respectively.
    RW and T denote the use of sample reweighting, and input transformation, respectively.
    }
    \vspace{-0.3cm}
    \label{fig:ablation_1}
\end{minipage}
\hfill
\begin{minipage}[t]{0.48\linewidth}
\centering
    \includegraphics[width=\linewidth]{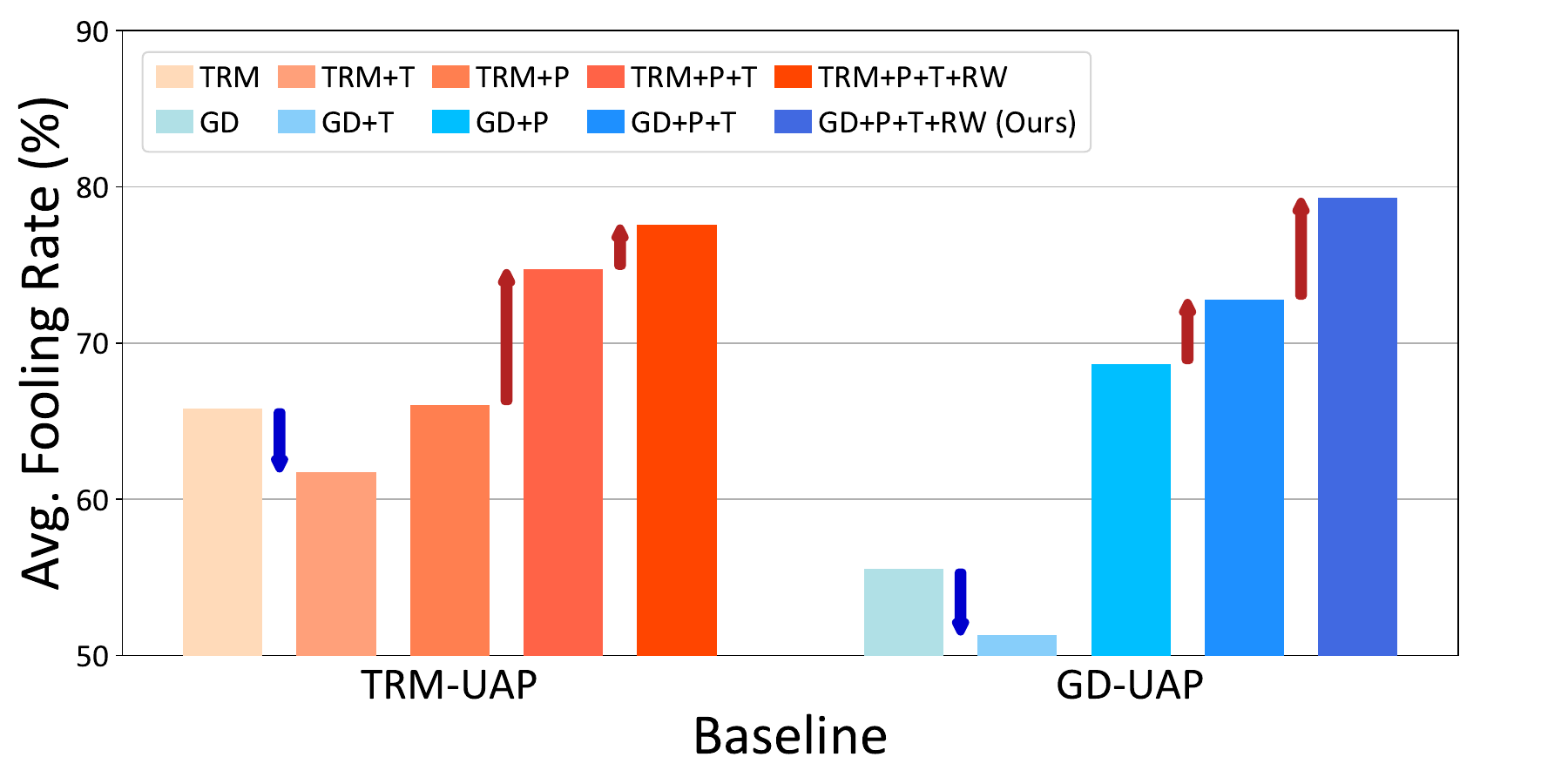}
    \vspace{-0.8cm}
    \caption{Demonstrating that PSP serves as a universal strategy to other data-free UAP methods. Avgerage fooling rate (\%) refers average FR (\%) on AlexNet, VGG16, VGG19, ResNet152, and GoogleNet, with UAP crafted on ResNet152. P, T, and RW denote PSP, input transformation, and sample reweighting, respectively.}
    
      
    \label{fig:ablation_2}
    \vspace{-0.3cm}
\end{minipage}
\end{figure*}

\vspace{-0.2cm}
\paragraph{Impact of Each Proposed Component.} 
We investigate how each component of PSP-UAP such as pseudo-semantic prior, sample reweighting, and input transformation affects attack performance.
For a fair comparison, we assign the number of random noises and semantic samples to 10.
As shown in Figure~\ref{fig:ablation_1}, our pseudo-semantic prior method achieves a higher $FR$ rather than random prior when used as an input prior.
Additionally, applying sample reweighting to the semantic samples improves the $FR$ in both white-box and black-box attacks.
Incorporating input transformation into semantic samples enhances the attack performance.
Remarkably, combining both sample reweighting and input transformation yields the greatest overall performance improvement. 

 

\vspace{-0.2cm}
\paragraph{Impact of Input Transformation.}
We evaluate the impact of applying input transformations to semantic samples in our method versus random noise on two data-free UAP methods (\eg, TRM-UAP~\cite{TRM} and GD-UAP~\cite{gd-uap}), to show that input transformations are effective when semantic information is present.
For a fair evaluation, we set the number of samples $N=10$ for both random noise and semantic samples.
As shown in Figure~\ref{fig:ablation_2}, applying input transformation directly to random noise leads to performance degradation, whereas integrating our pseudo-semantic prior (PSP) significantly boosts performance. 
Moreover, each baseline achieves its highest $FR$ when combined with our full approach, including sample reweighting.
We believe that this improvement results from the richer semantic content provided by our PSP, which allows the UAP to learn a broader range of patterns through input transformation.
This experiment also validates our approach as a universal strategy.

\begin{figure}[t]
\centering
    \includegraphics[width=\linewidth]{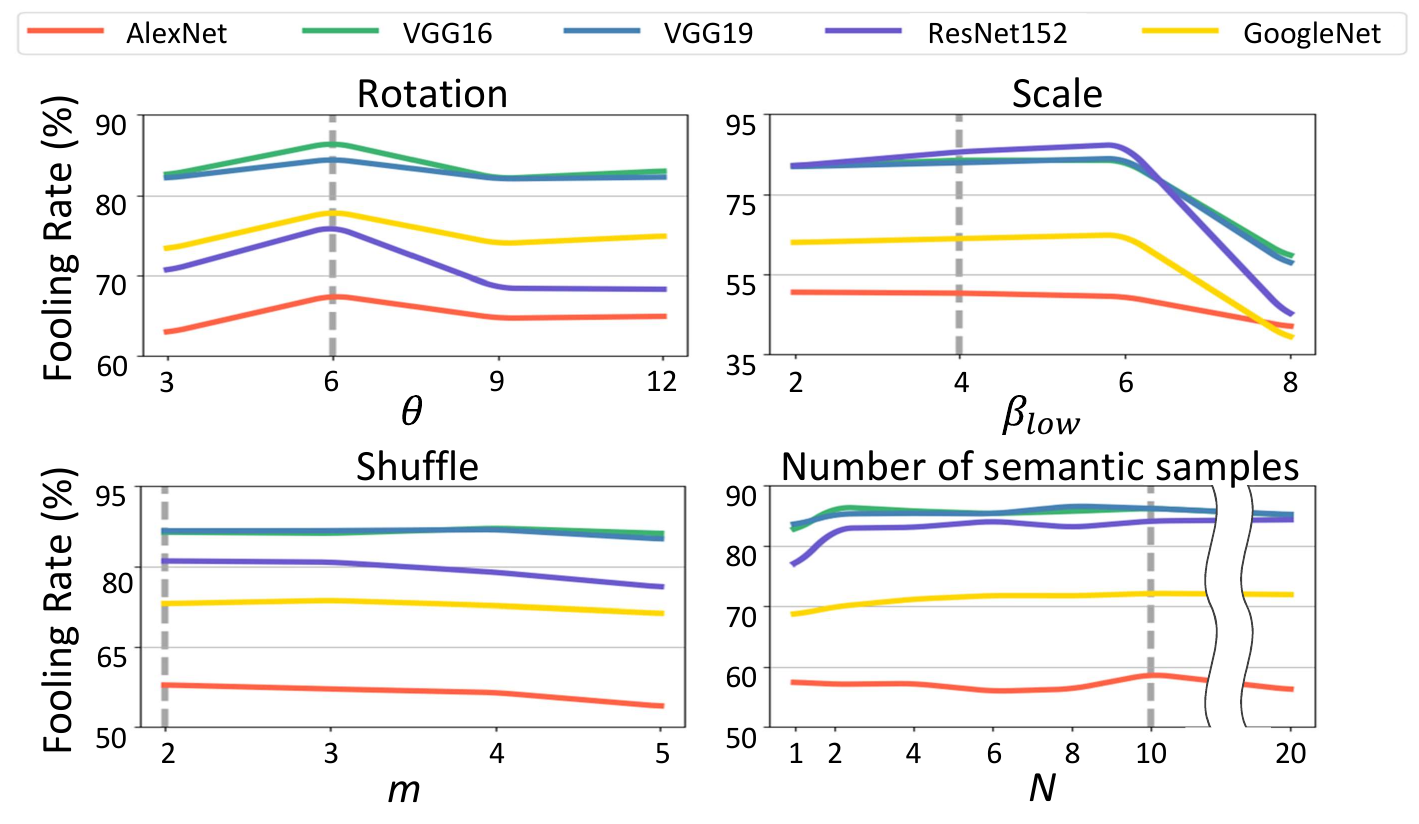}
    \vspace{-0.5cm}
    \caption{
    Ablation study on the hyperparameters for input transformation and the number of semantic samples. The hyperparameters used in our experiments are marked with gray dashed line. 
    }
    \label{fig:ablation_num}
    \vspace{-0.2cm}
\end{figure}

\vspace{-0.2cm}
\paragraph{Hyperparameter Analysis.}
We conduct a hyperparameter analysis to evaluate how different settings for input transformation and the number of semantic samples affect our method's attack performance on the ImageNet validation set (see Figure~\ref{fig:ablation_num}).
With the exception of extreme scaling values (\eg, 8), the proposed method demonstrates stable performance across various hyperparameter values.
Furthermore, performance remains consistent across different the number of semantic samples; notably, even with $N=1$, method outperforms other data-free approaches.
Note that all hyperparameters used are chosen based on the ImageNet train set (See Appendix for further details).

\begin{figure}[t]
\centering
    \includegraphics[width=1\linewidth]{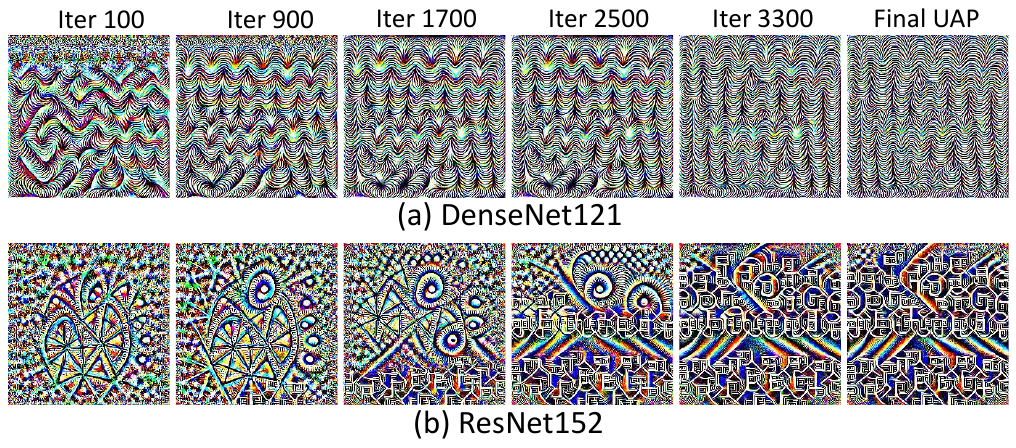}
    \vspace{-0.6cm}
    \caption{Visualization of UAPs crafted by PSP-UAP during training on DenseNet121 and ResNet152. From left to right, the UAPs are shown at iterations 100, 900, 1700, 2500, 3300, and the final UAP after training. Pixel values are scaled to [0, 255].
  }
  \vspace{-0.2cm}
    \label{fig:uaps}
\end{figure}

\paragraph{UAP Visualization.}
In Figure~\ref{fig:uaps}, we visualize the UAPs at each training iteration to verify that the pseudo-semantic prior, constructed from UAPs in the training phase, captures a variety of inherent patterns.
We observe that the UAPs contain more diverse patterns in the early stages of training.
We believe this variability allows us to obtain semantic samples with a broad range of patterns, through which the UAP learns diverse semantic representations.

%% file: table/black_box_alex.tex
\begin{table*}[t]
\footnotesize
  \centering
  \begin{tabular}{cccccccc}
    \toprule
    Model & Attack & AlexNet~~ & VGG16~~ & VGG19~~ & ResNet152~~ & GoogleNet~~ & Average\\
    \midrule
    \multirow{3}{*}{AlexNet} 
    & AT-UAP-U & \textbf{96.66}\scriptsize{$\pm$0.12}* & {72.33}\scriptsize{$\pm$0.50}~~ & {67.24}\scriptsize{$\pm$0.18}~~ & {43.63}\scriptsize{$\pm$0.29}~~ & {62.01}\scriptsize{$\pm$0.32}~~ & {68.37}\\
    & TRM-UAP & 93.53\scriptsize{$\pm$0.07}* & 60.10\scriptsize{$\pm$0.24}~~ & 57.08\scriptsize{$\pm$0.15}~~ & 27.31\scriptsize{$\pm$0.30}~~ & 32.70\scriptsize{$\pm$0.22}~~ & 54.14 \\
    & \textbf{PSP-UAP (Ours)} & 91.77\scriptsize{$\pm$0.32}* & \textbf{76.56}\scriptsize{$\pm$0.67}~~ & \textbf{74.07}\scriptsize{$\pm$0.54}~~ & \textbf{49.20}\scriptsize{$\pm$1.12}~~ & \textbf{66.00}\scriptsize{$\pm$0.75}~~ & \textbf{71.52 }\\
    \hline
    
    \multirow{3}{*}{VGG16} 
    & AT-UAP-U & \textbf{54.15}\scriptsize{$\pm$0.70}~~ & 94.50\scriptsize{$\pm$0.21}* & 86.65\scriptsize{$\pm$0.70}~~ & 36.96\scriptsize{$\pm$1.03}~~ & 48.53\scriptsize{$\pm$1.32}~~ & 64.16 \\
    & TRM-UAP & 47.53\scriptsize{$\pm$0.51}~~ & 94.30\scriptsize{$\pm$0.12}* & 89.68\scriptsize{$\pm$0.14}~~ & {61.43}\scriptsize{$\pm$0.40}~~ & 53.95\scriptsize{$\pm$0.59}~~ & {69.38}\\
    & \textbf{PSP-UAP (Ours)} & {50.40}\scriptsize{$\pm$0.53}~~ & \textbf{96.26}\scriptsize{$\pm$0.21}* & \textbf{92.60}\scriptsize{$\pm$0.33}~~ & \textbf{74.10}\scriptsize{$\pm$1.10}~~ & \textbf{64.89}\scriptsize{$\pm$0.66}~~ & \textbf{75.65}\\
    \hline
    
    \multirow{3}{*}{VGG19} 
    & AT-UAP-U & \textbf{62.05}\scriptsize{$\pm$1.01}~~ & 88.96\scriptsize{$\pm$0.50}~~ & 92.85\scriptsize{$\pm$0.48*} & 42.72\scriptsize{$\pm$0.51}~~ & \textbf{60.99}\scriptsize{$\pm$1.41}~~ & 69.51\\
    & TRM-UAP & 46.01\scriptsize{$\pm$0.44}~~ & 89.82\scriptsize{$\pm$0.15}~~ & 91.35\scriptsize{$\pm$0.30}* & 47.19\scriptsize{$\pm$0.46}~~ & 46.48\scriptsize{$\pm$0.78}~~ & 64.17\\
    & \textbf{PSP-UAP (Ours)} & 48.93\scriptsize{$\pm$0.72}~~ & \textbf{94.55}\scriptsize{$\pm$0.14}~~ & \textbf{94.65}\scriptsize{$\pm$0.10}* & \textbf{67.13}\scriptsize{$\pm$1.37}~~ & {58.83\scriptsize{$\pm$1.19}}~~ & \textbf{72.81}\\
    \hline
    
    \multirow{3}{*}{ResNet152} 
    & AT-UAP-U & 49.78\scriptsize{$\pm$0.68}~~ & 62.78\scriptsize{$\pm$0.71}~~ & 60.54\scriptsize{$\pm$0.49}~~ & 73.15\scriptsize{$\pm$1.15}* & 48.37\scriptsize{$\pm$0.49}~~ & 58.92\\
    & TRM-UAP & {53.56}\scriptsize{$\pm$0.75}~~ & 77.20\scriptsize{$\pm$0.35}~~ & 73.30\scriptsize{$\pm$0.41}~~ & 67.46\scriptsize{$\pm$0.35}* & 57.54\scriptsize{$\pm$0.50}~~ & 65.81\\
    & \textbf{PSP-UAP (Ours)} & \textbf{58.82}\scriptsize{$\pm$1.17}~~ & \textbf{88.59}\scriptsize{$\pm$1.38}~~ & \textbf{87.35}\scriptsize{$\pm$0.92}~~ & \textbf{85.65}\scriptsize{$\pm$1.70}* & \textbf{76.00}\scriptsize{$\pm$1.33}~~ & \textbf{79.28 }\\
    \hline
    
    \multirow{3}{*}{GoogleNet} 
    & AT-UAP-U & 55.65\scriptsize{$\pm$0.37}~~ & 71.38\scriptsize{$\pm$0.83}~~ & 68.25\scriptsize{$\pm$0.59}~~ & 43.03\scriptsize{$\pm$0.42}~~ & 82.60\scriptsize{$\pm$0.72}* & 64.18\\
    & TRM-UAP & 60.10\scriptsize{$\pm$1.16}~~ & \textbf{79.66}\scriptsize{$\pm$0.95}~~ & \textbf{79.98}\scriptsize{$\pm$1.06}~~ & \textbf{58.85}\scriptsize{$\pm$1.94}~~ & \textbf{85.32}\scriptsize{$\pm$0.04}* & \textbf{72.78}\\
    & \textbf{PSP-UAP (Ours)} & \textbf{65.22}\scriptsize{$\pm$0.56}~~ & 78.43\scriptsize{$\pm$0.73}~~ & 79.26\scriptsize{$\pm$0.73}~~ & 57.63\scriptsize{$\pm$0.66}~~ & 81.43\scriptsize{$\pm$0.49}* & 72.39\\
    \bottomrule
  \end{tabular}
  \caption{Black-box attack transferability of the UAP synthesized by our PSP-UAP method compared to other data-free universal attacks, AT-UAP-U~\cite{at-uap} and TRM-UAP~\cite{TRM}. We show the mean and standard deviation of $FR$ with five runs. Bold $FR$ (\%) denotes the best performance. The UAPs are crafted on AlexNet, VGG16, VGG19, ResNet152, and GoogleNet. * indicate $FR$ of the white-box model. }
  \label{tab:black_box_combined}
\end{table*}

%% file: table/black_box.tex
\begin{table*}
\footnotesize
  \centering
  \begin{tabular}{ccccccc}
    \toprule
    Model & Attack & ResNet50 & DenseNet121 & MobileNet-v3-L & Inception-v3 & Average\\
    \midrule
    \multirow{2}{*}{ResNet50} & TRM-UAP & 73.26\scriptsize{$\pm$0.82}* & 54.42\scriptsize{$\pm$1.23}~~ &  61.25\scriptsize{$\pm$1.48}~~ &  37.36\scriptsize{$\pm$0.69}~~ & 56.57\\
    & \textbf{PSP-UAP (Ours)} & \textbf{77.60}\scriptsize{$\pm$0.42}* & \textbf{66.11}\scriptsize{$\pm$0.87}~~ & \textbf{70.50}\scriptsize{$\pm$1.10}~~ &  \textbf{42.32}\scriptsize{$\pm$1.32}~~ & \textbf{64.13}\\
    \hline
    \multirow{2}{*}{DenseNet121} & TRM-UAP & 35.24\scriptsize{$\pm$2.55}~~ & 70.10\scriptsize{$\pm$2.07}* &  34.17\scriptsize{$\pm$1.77}~~ &  32.11\scriptsize{$\pm$2.38}~~ & 42.91\\
    & \textbf{PSP-UAP (Ours)} & \textbf{53.03}\scriptsize{$\pm$0.90}~~ & \textbf{85.81}\scriptsize{$\pm$1.17}* & \textbf{50.22}\scriptsize{$\pm$0.58}~~ & \textbf{50.73}\scriptsize{$\pm$0.78}~~ & \textbf{59.95}\\
    \hline
    \multirow{2}{*}{MobileNet-v3-L} & TRM-UAP & 39.47\scriptsize{$\pm$1.11}~~ & 40.37\scriptsize{$\pm$0.47}~~ &  73.07\scriptsize{$\pm$0.96}* & 30.11\scriptsize{$\pm$0.81}~~ & 45.76 \\
    & \textbf{PSP-UAP (Ours)} & \textbf{54.38}\scriptsize{$\pm$1.40}~~ & \textbf{54.62}\scriptsize{$\pm$1.82}~~ &  \textbf{90.39}\scriptsize{$\pm$0.23}* &  \textbf{46.29}\scriptsize{$\pm$0.69}~~ & \textbf{61.42}\\
    \hline
     \multirow{2}{*}{Inception-v3} & TRM-UAP & 53.53\scriptsize{$\pm$0.57}~~ & 54.93\scriptsize{$\pm$0.54}~~ &  67.16\scriptsize{$\pm$0.60}~~ & 64.22\scriptsize{$\pm$0.33}* &59.96\\
    & \textbf{PSP-UAP (Ours)} & \textbf{57.60}\scriptsize{$\pm$0.26}~~  & \textbf{57.50}\scriptsize{$\pm$0.59}~~ & \textbf{70.20}\scriptsize{$\pm$0.56}~~ & \textbf{65.38}\scriptsize{$\pm$0.51}* & \textbf{62.67}\\
    \bottomrule
  \end{tabular}
  \caption{$FR$ (\%) for the UAPs crafted by TRM-UAP~\cite{TRM} and our PSP-UAP across additional CNN models. The UAPs are crafted on ResNet50, DenseNet121, MobileNet-v3-Large, and Inception-v3. * indicates the white-box model.}
  \vspace{-0.2cm}
  \label{tab:black_box_advance}
\end{table*}


%% file: table/data-dependent.tex
\begin{table*}[t]
\footnotesize
  \centering
  \begin{tabular}{ccccccccc}
    \toprule
    Model & Data & Attack & AlexNet~~ & VGG16~~ & VGG19~~ & ResNet152~~ & GoogleNet~~ & Average\\
    \midrule
    \multirow{3}{*}{AlexNet} 
    & \multirow{2}{*}{\textcolor[rgb]{0,0.7,0}{\ding{51}}} & SGA-UAP & \textbf{97.43}* & 66.41 & 60.96 & 35.76 & 49.71 & 62.05 \\
    & & AT-UAP-S & 97.01\scriptsize{$\pm$0.11}* & 62.37\scriptsize{$\pm$1.37}~~ & 57.72\scriptsize{$\pm$0.62}~~ & 33.40\scriptsize{$\pm$0.77}~~ & 47.31\scriptsize{$\pm$1.65}~~ & 59.56 \\
    \cdashline{2-9}[1pt/1pt]
    & \textcolor[rgb]{0.9,0,0}{\ding{55}} & \textbf{PSP-UAP (Ours)} & 91.77\scriptsize{$\pm$0.32}* & \textbf{76.56}\scriptsize{$\pm$0.67}~~ & \textbf{74.07}\scriptsize{$\pm$0.54}~~ & \textbf{49.20}\scriptsize{$\pm$1.12}~~ & \textbf{66.00}\scriptsize{$\pm$0.75}~~ & \textbf{71.52 }\\
    \hline
    
    \multirow{3}{*}{VGG16} 
    & \multirow{2}{*}{\textcolor[rgb]{0,0.7,0}{\ding{51}}} & SGA-UAP & 49.02 & \textbf{98.36}* & \textbf{94.17} & 49.02 & {55.78} & 69.27 \\
    & & AT-UAP-S  & 45.58\scriptsize{$\pm$0.29}~~ & 97.51\scriptsize{$\pm$0.08}* & 91.53\scriptsize{$\pm$0.22}~~ & 47.16\scriptsize{$\pm$0.95}~~ & 53.63\scriptsize{$\pm$0.90}~~ & 67.08\\
    \cdashline{2-9}[1pt/1pt]
    & \textcolor[rgb]{0.9,0,0}{\ding{55}} & \textbf{PSP-UAP (Ours)} & \textbf{50.40}\scriptsize{$\pm$0.53}~~ & {96.26}\scriptsize{$\pm$0.21}* & {92.60}\scriptsize{$\pm$0.33}~~ & \textbf{74.10}\scriptsize{$\pm$1.10}~~ & \textbf{64.89}\scriptsize{$\pm$0.66}~~ & \textbf{75.65}\\
    \hline
    
    \multirow{3}{*}{VGG19} 
    & \multirow{2}{*}{\textcolor[rgb]{0,0.7,0}{\ding{51}}} & SGA-UAP& \textbf{50.67} & \textbf{95.52} & \textbf{97.69}* & {51.08} & 56.87 & {70.37}\\
    & & AT-UAP-S & 46.04\scriptsize{$\pm$0.58}~~ & 93.49\scriptsize{$\pm$0.17}~~ & 97.56\scriptsize{$\pm$0.04}* & 43.53\scriptsize{$\pm$0.57}~~ & 52.58\scriptsize{$\pm$0.81}~~ & 66.64\\
    \cdashline{2-9}[1pt/1pt]
    & \textcolor[rgb]{0.9,0,0}{\ding{55}} & \textbf{PSP-UAP (Ours)} & 48.93\scriptsize{$\pm$0.72}~~ & {94.55}\scriptsize{$\pm$0.14}~~ & {94.65}\scriptsize{$\pm$0.10}* & \textbf{67.13}\scriptsize{$\pm$1.37}~~ & \textbf{58.83}\scriptsize{$\pm$1.19}~~ & \textbf{72.81}\\
    \hline
    
    \multirow{3}{*}{ResNet152} 
    & \multirow{2}{*}{\textcolor[rgb]{0,0.7,0}{\ding{51}}} & SGA-UAP & 51.59 & {81.77} & {79.01} & \textbf{94.04}* & {64.05} & {74.09}\\
    & & AT-UAP-S & 47.33\scriptsize{$\pm$0.89}~~ & 81.93\scriptsize{$\pm$0.94}~~ & 78.72\scriptsize{$\pm$0.91}~~ & 91.52\scriptsize{$\pm$0.78}* & 61.32\scriptsize{$\pm$0.98}~~ & 72.16\\
    \cdashline{2-9}[1pt/1pt]
    & \textcolor[rgb]{0.9,0,0}{\ding{55}} & \textbf{PSP-UAP (Ours)} & \textbf{58.82}\scriptsize{$\pm$1.17}~~ & \textbf{88.59}\scriptsize{$\pm$1.38}~~ & \textbf{87.35}\scriptsize{$\pm$0.92}~~ & 85.65\scriptsize{$\pm$1.70}* & \textbf{76.00}\scriptsize{$\pm$1.33}~~ & \textbf{79.28 }\\
    \hline
    
    \multirow{3}{*}{GoogleNet} 
    & \multirow{2}{*}{\textcolor[rgb]{0,0.7,0}{\ding{51}}} & SGA-UAP & {62.56} & \textbf{83.62} & \textbf{82.11} & \textbf{59.09} & \textbf{92.12}* & \textbf{75.90}\\
    & & AT-UAP-S & 55.90\scriptsize{$\pm$0.62}~~ & 78.71\scriptsize{$\pm$0.67}~~ & 76.01\scriptsize{$\pm$0.45}~~ & 54.49\scriptsize{$\pm$0.29}~~ & 90.82\scriptsize{$\pm$0.29}* & 71.19 \\
    \cdashline{2-9}[1pt/1pt]
    & \textcolor[rgb]{0.9,0,0}{\ding{55}} & \textbf{PSP-UAP (Ours)} & \textbf{65.22}\scriptsize{$\pm$0.56}~~ & 78.43\scriptsize{$\pm$0.73}~~ & 79.26\scriptsize{$\pm$0.73}~~ & 57.63\scriptsize{$\pm$0.66}~~ & 81.43\scriptsize{$\pm$0.49}* & 72.39 \\
    \bottomrule
  \end{tabular}
  \caption{$FR$ (\%) of our PSP-UAP and data-dependent UAPs. SGA-UAP~\cite{sga-uap} and AT-UAP-S~\cite{at-uap}. The "data" column indicates whether a dataset was used to train UAPs (data-dependent UAP, \textcolor[rgb]{0,0.7,0}{\ding{51}}) or not (data-free UAP, \textcolor[rgb]{0.9,0,0}{\ding{55}}). * indicates the white-box model.
  }
  \vspace{-0.2cm}
  \label{tab:black_box_data_dep}
\end{table*}

%% file: sec/5_Conclusion.tex
\section{Conclusion}

In this paper, we proposed a novel data-free universal attack method that leverages UAPs as a prior enriched with semantic information.
This approach allows us to directly draw semantic samples from the pseudo-semantic prior, overcoming the lack of target domain knowledge.
To further enhance transferability, we applied input transformation methods to these semantic samples.
Additionally, we introduced sample reweighting to ensure a balanced attack across semantic samples.
We demonstrated the exceptional transferability of our method by comparing PSP-UAP with both data-free and data-dependent universal attack approaches across various CNN models on the ImageNet validation dataset.

\paragraph{Acknowledgements.}
\small
This work was supported by the IITP grants (RS-2019-II191842~(3\%), RS-2021-II212068~(2\%), RS-2022-II220926~(50\%)) funded by MSIT, and the GIST-MIT Research Collaboration grant~(45\%) funded by GIST, Korea.

%% file: sec/X_suppl.tex
\clearpage
\appendix
\setcounter{page}{1}
\maketitle
%
%
%

\section{Additional Experiments on CNN Models}

\begin{figure}[t]
\centering
    \includegraphics[width=1\linewidth]{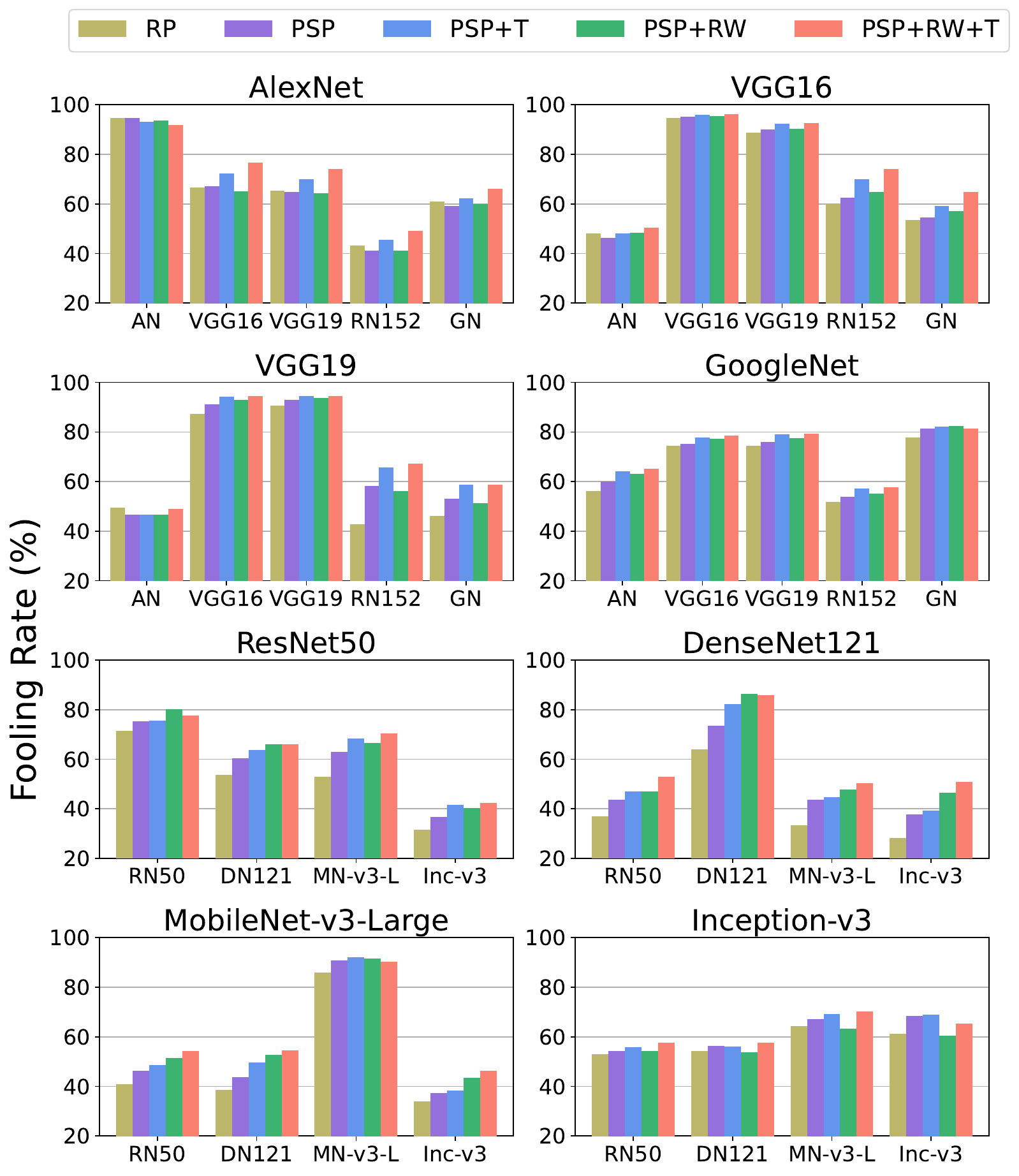}
    \vspace{-0.5cm}
    \caption{Ablation study on each proposed component in PSP-UAP on various CNN models. RP and PSP refer to training a UAP using random noises and semantic samples drawn from pseudo-semantic prior.
    RW and T denote the use of sample reweighting, and input transformation, respectively.
    All experiments, including, RP are conducted with the number of samples, $N$, set to $10$.
  }
   \vspace{-0.5cm}
    \label{fig:sup_all}
\end{figure}

\paragraph{Impact of Each Proposed Component.}
In the main manuscript, we evaluate the impact of each component on attack performance by generating UAPs using ResNet152.
In this section, we extend our ablation study to other models, with the results summarized in Figure~\ref{fig:sup_all}.
AlexNet is abbreviated as AN, ResNet152 as RN152, GoogleNet as GN, ResNet50 as RN50, DenseNet121 as DN121, MobileNet-v3-Large as MN-v3, and Inception-v3 as Inc-v3.
We observe consistent trends across models, where the incorporation of pseudo-semantic priors (PSP), sample reweighting, and input transformation enhances the attack performance of the generated UAPs. 
However, the effect of PSP is less pronounced in AlexNet, while input transformations have a reduced impact on white-box attack performance.
Additionally, in experiments with VGG19 and Inception-v3, several models demonstrate reduced performance when sample reweighting is applied alone.
Despite these minor degradations, our full model, which combines all components, achieves a significantly higher black-box fooling rate on average, demonstrating its robustness even when individual components show limited effectiveness.

\paragraph{Additional Experiments for Transferability.}
We conduct additional experiments to explore the black-box attack transferability across various models further.
We generate UAPs on ResNet50, DenseNet121, MobileNet-v3-Large, and Inception-v3, and attack AlexNet, VGG16, VGG19, ResNet152, and GoogleNet.
The results in Table~\ref{tab:add_to_trad} demonstrate that our method persistently surpasses TRM-UAP in attack performance, even when the target model changes, highlighting its superiority.



\paragraph{Impact of Epsilon.}

We evaluate the impact of $\epsilon$ which is a constraint parameter that restricts the pixel intensity of the generated UAPs used in Eq.~(8) of the main manuscript. 
Note that, for experiments in the main manuscript, we set $\epsilon=10$, following the conventional setting of data-free UAP methods.
To further analyze its effect, we compare the $FR$ of our method with TRM-UAP using various $\epsilon$ values of 8, 10, and 16.
The results, shown in Table~\ref{tab:black_box_epsilonj}, show that our method consistently outperforms TRM-UAP in terms of $FR$ across different values of $\epsilon$.
These experiments demonstrate that the pseudo-semantic prior retains sufficient value as the data prior, even under varying levels of constraints.

\input{table/add_to_trad}

\input{table/supple_epsilon}

\paragraph{Robustness against Defenses.}
In Table~\ref{tab:jpeg}, we validate the robustness of our method against JPEG compression~\cite{jpeg} and ensemble adversarially trained models, ens3-adv-Inc-v3 and ens-adv-Inc-Res-v2~\cite{ens_adv}.
Our method consistently shows higher robustness than TRM-UAP, with UAP crafted on ResNet152.

\input{table/jpeg}

\input{table/other_models}

\paragraph{Diverse Surrogate Models}
We craft UAPs on various models in the main manuscript.
To further demonstrate the effectiveness of our method on recent architectures, we additionally evaluate it using ConvNext-B~\cite{convnext} and DeiT3~\cite{deit} as surrogate models.
As shown in Table~\ref{tab:convnext_vit}, our method achieves strong performance on ConvNext-B but underperforms on DeiT3, which is consistent with the behavior observed in TRM.
We attribute this to the fact that both our baseline and TRM were originally designed for CNN-based models, which may result in limited effectiveness on ViT-based architectures.



\begin{figure*}[t]
\centering
    \includegraphics[width=0.95\linewidth]{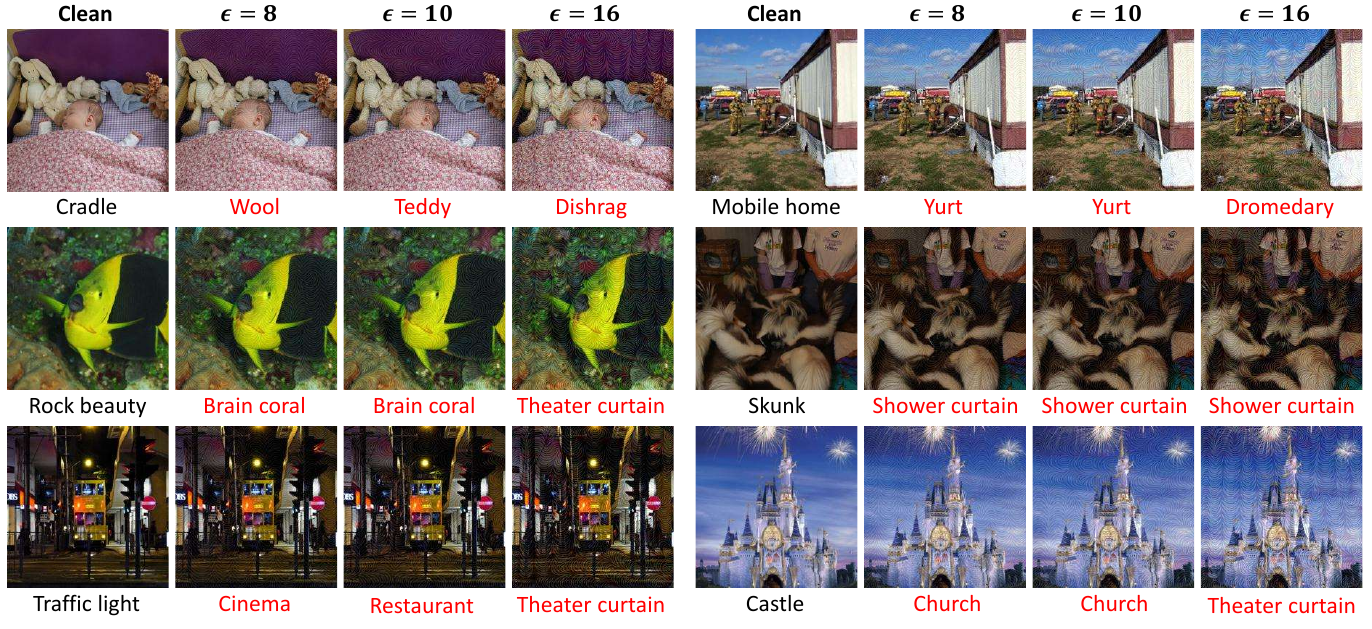}
    \caption{Qualitative results of our method. The leftmost column represents the original images, while the remaining three columns correspond to adversarial images generated with $\epsilon = 8$, $10$, and $16$ (from left to right). The predicted labels are displayed below each image. The UAPs are crafted on ResNet152.
  }
    \label{fig:sup_quali}
\end{figure*}

\begin{figure*}[t]
\centering
    \includegraphics[width=0.95\linewidth]{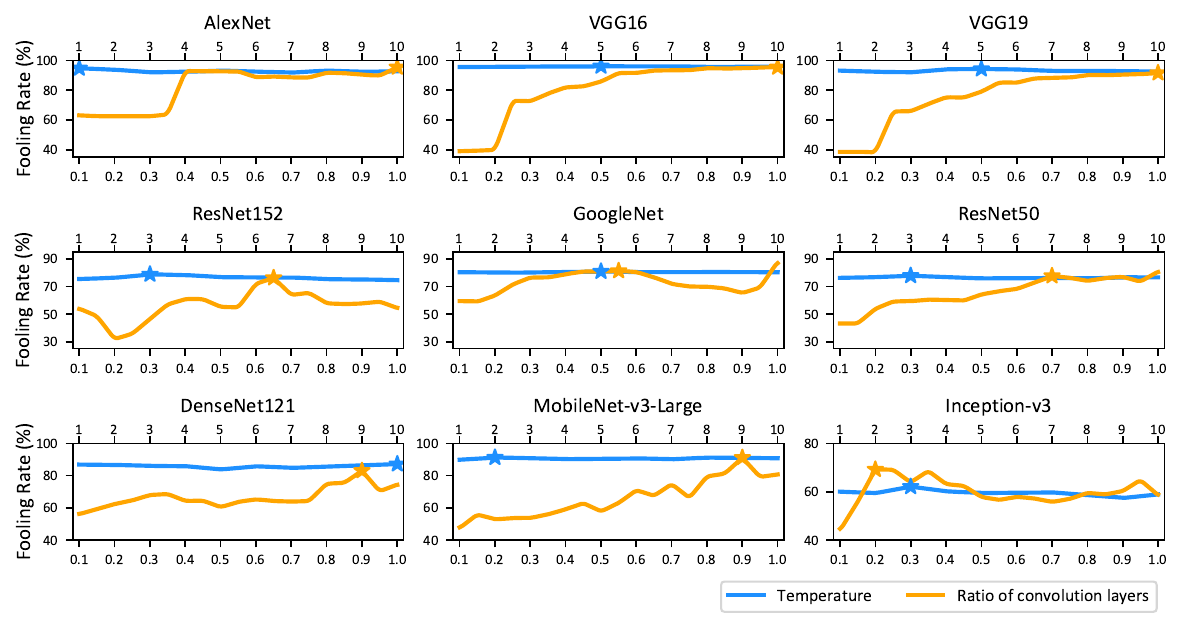}
    \caption{Parameter study on the ratio of convolutional layers and the temperature parameter for sample reweighting.
  }
    \label{fig:sup_hyper}
\end{figure*}

\paragraph{Qualititive Results}
We illustrate adversarial examples attacked by our generated UAPs using ResNet152 in Figure~\ref{fig:sup_quali} with different $\epsilon\in\{8,10,16\}$. 
As expected, smaller $\epsilon$ values result in minimal degradation to the original image, whereas larger $\epsilon$ values highlight more artifacts introduced by the UAP.
Similarly, as shown in Table~\ref{tab:black_box_epsilonj}, smaller $\epsilon$ values lead to lower performance compared to larger $\epsilon$ values.
We also visualize the final UAPs crafted for each model and the intermediate UAPs used during the training phase to generate the pseudo-semantic prior in Figure~\ref{fig:sup_uap_1}.
As discussed in the main manuscript, visually diverse patterns can be observed across different iterations, even on the same surrogate model.
This demonstrates that our method effectively crafts UAPs even in the absence of prior knowledge by generating diverse semantic samples.

\section{Ablation Study on Hyperparameters}
\input{table/temper_tradcnn}

\input{table/temper_addcnn}
In this section, we demonstrate an ablation study on the hyperparameters used in our PSP-UAP framework, including the ratios of convolutional layers to calculate the loss, temperature parameters in the sample reweighting, and the ranges for rotation, scaling, and shuffling in the input transformation.
To determine the optimal set of parameters, we follow the setting used in previous works~\cite{gd-uap, TRM}.

\paragraph{Ratio of Convolutional Layers}
We follow the same process outlined in TRM-UAP~\cite{TRM} to determine $l'$ in Eq.~(8) by searching for the optimal ratio of convolutional layers.
For this, we use only our pseudo-semantic priors, excluding sample reweighting and input transformation.
Figure~\ref{fig:sup_hyper} shows the results, with yellow lines indicating outcomes and the yellow star marking the convolutional layer ratios used in our experiments.
Based on this, the ratios are set to $100\%$, $100\%$, $100\%$, $65\%$, $55\%$, $70\%$, $90\%$, $90\%$, $20\%$ for AlexNet, VGG16, VGG19, ResNet152, GoogleNet, ResNet50, DenseNet121, MobileNet-v3-Large, Inception-v3, respectively.
Note that, for a fair comparison with TRM-UAP in Table 3 of our main manuscript and Table 5 in this supplementary material, we made every effort to conduct comprehensive experiments to determine the optimal positive truncation rate (PTR) and negative truncation rate (NTR) for TRM-UAP.

\paragraph{Temperature Parameters}

After determining the optimal convolution layer ratio, we use it as a basis to find the temperature parameter $\tau$, used in Eq.~(6) for the temperature-scaled softmax output, by incrementally increasing it from 1 to 10 in steps of 1.
The results are shown in Figure~\ref{fig:sup_hyper}, with blue lines representing the outcomes and the blue stars indicating the temperature values used in our experiments.
Our observations indicate that variations in the temperature parameter $\tau$ have minimal impact on the results.
In Table~\ref{tab:temper_tradcnn} and Table~\ref{tab:black_box_temp_2}, we report the performances of our PSP-UAP with a fixed temperature ($\tau=4$, referred to as PSP-I) alongside PSP-UAP with optimal temperature values (PSP-D) and TRM-UAP for comparison.
Even with a fixed temperature, the performance difference is minimal, and our method consistently outperforms TRM-UAP by a significant margin. 
This highlights the robustness of our approach, achieving strong results over TRM-UAP even without tuning the temperature parameter.

\begin{figure}[t]
\centering
    \includegraphics[width=\linewidth]{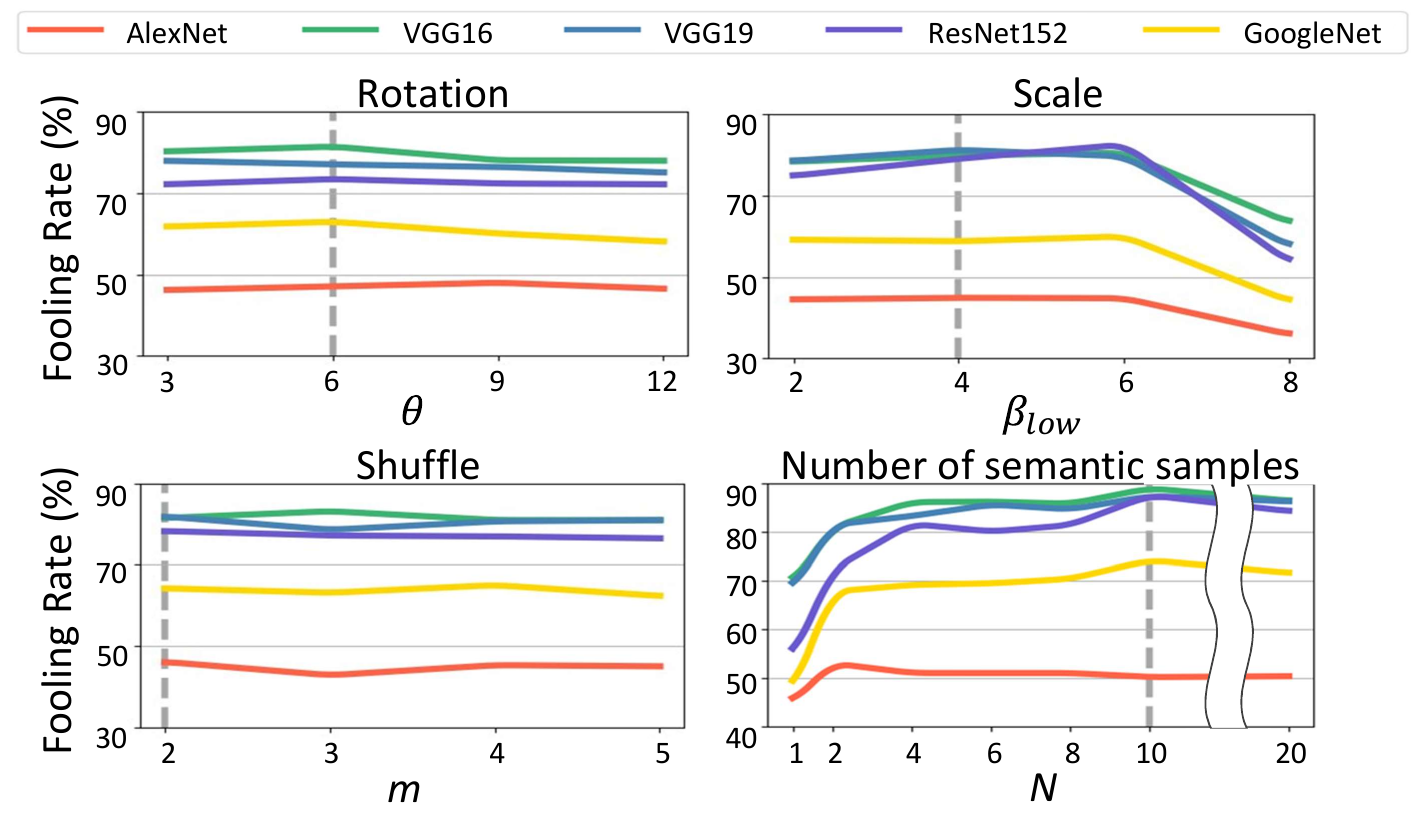}
    \vspace{-0.5cm}
    \caption{
    Hyperparameter analysis on ImageNet train set for input transformation and the number of semantic samples. The hyperparameters used in our experiments are marked with gray dashed line. 
    }
    \label{fig:hyper_train}
    \vspace{-0.2cm}
\end{figure}

\begin{figure*}[t]
\centering
    \includegraphics[width=0.95\linewidth]{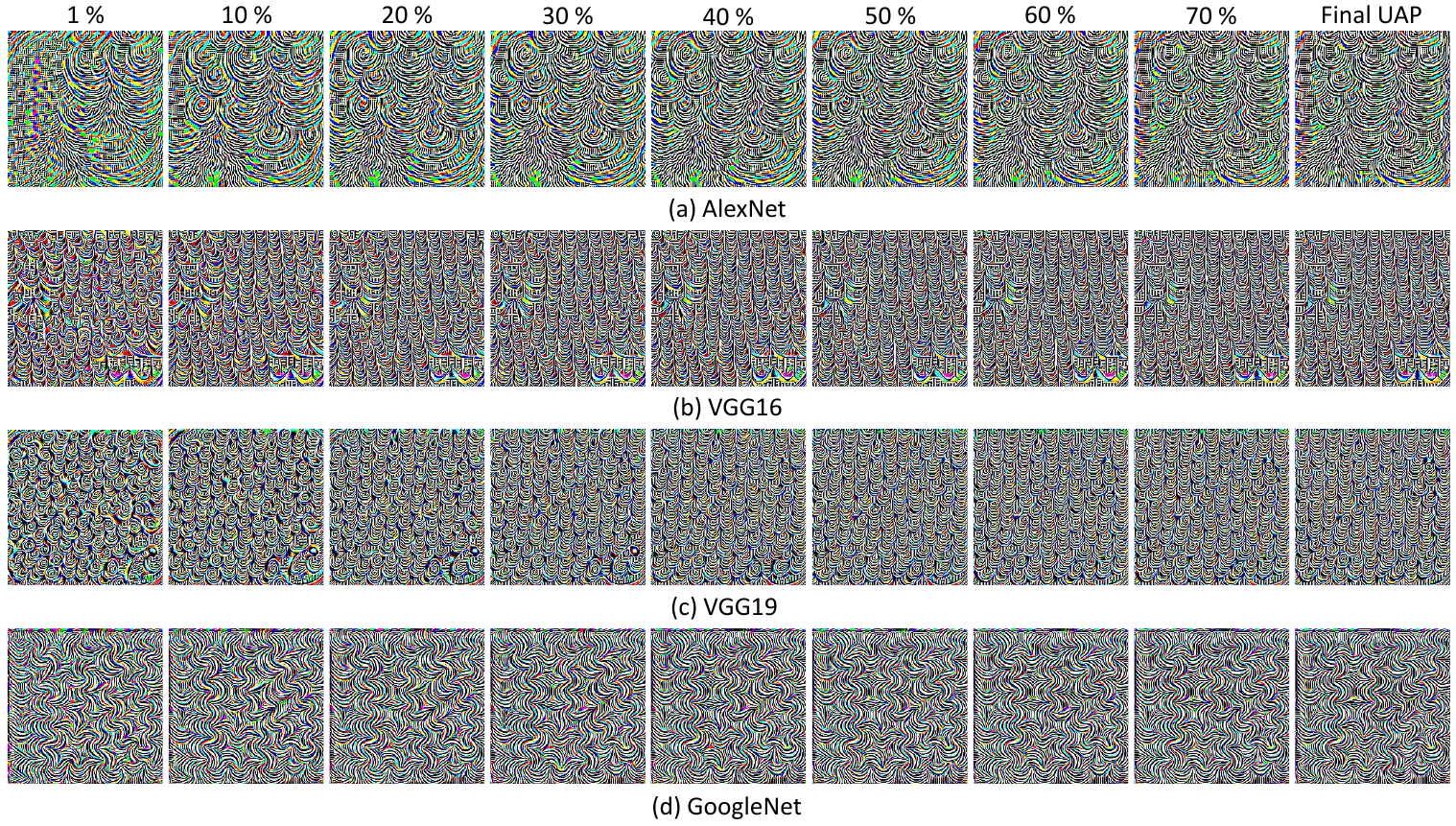}
    \includegraphics[width=0.95\linewidth]{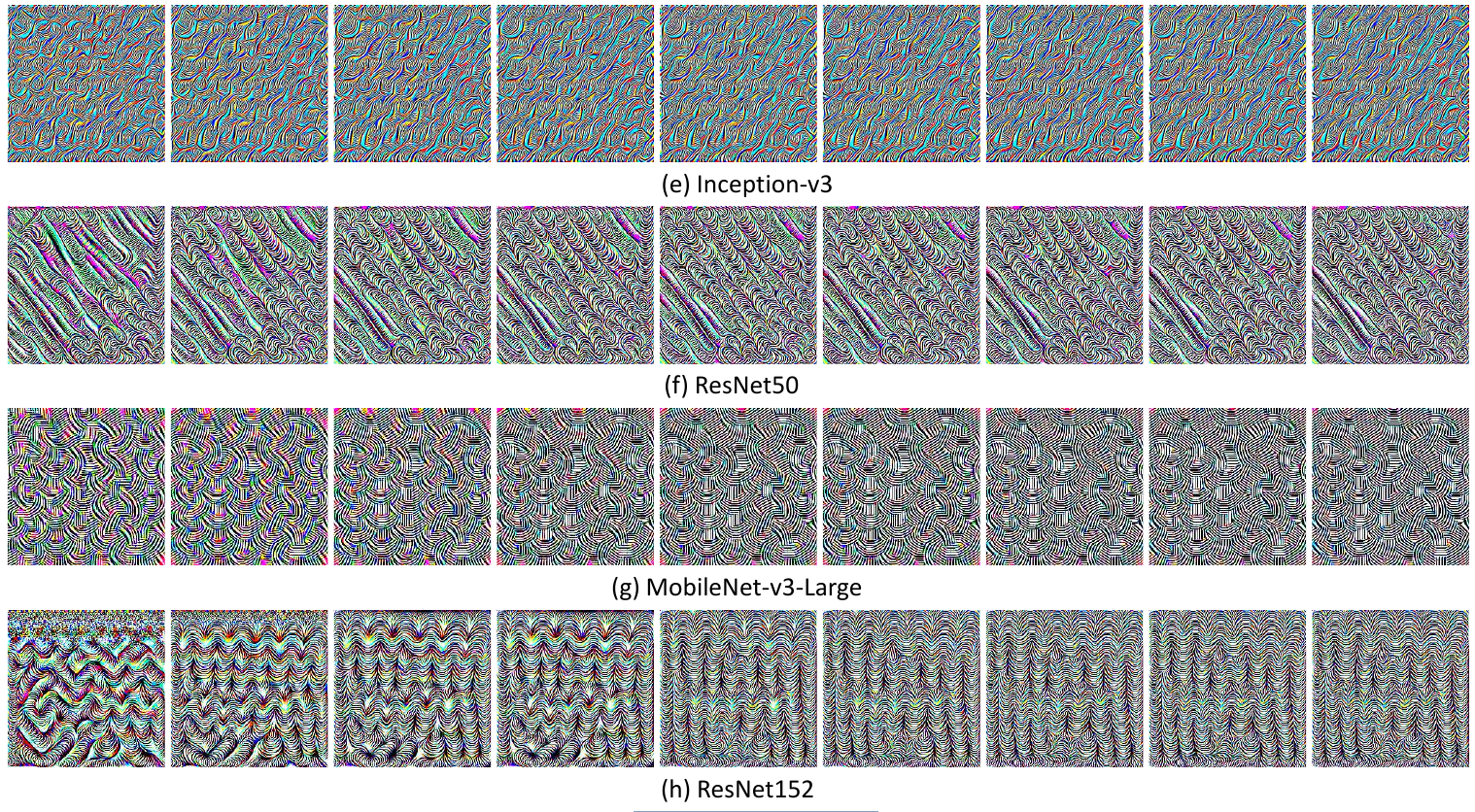}
    \includegraphics[width=0.95\linewidth]{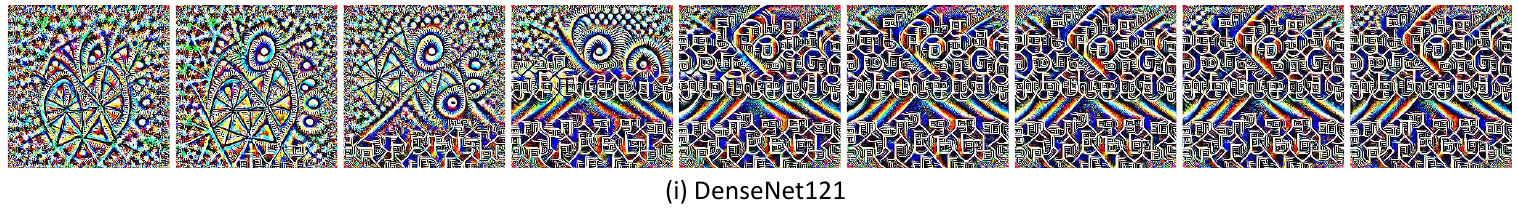}
    \caption{Visualization of the UAPs crafted by various CNN models in the training phase.
    The percentage above the figure corresponds to the progress of training iterations (\textit{e.g.,} $1000$ iterations out of $10000$ = $10\%$)
  }
    \label{fig:sup_uap_1}
\end{figure*}

\paragraph{Hyperparameter Search on ImageNet Train Set} 
We conduct experiments on a randomly selected subset of 1,000 images from the ImageNet train set to determine hyperparameters for input transformation and the number of semantic samples. 
As shown in Figure~\ref{fig:hyper_train}, our method demonstrates consistent performance across various transformation hyperparameters and exhibits a similar tendency to the results in Figure~6 of the main manuscript.
In the case of the number of semantic samples, the performance is relatively low when $N=1$, which is likely due to sample imbalance caused by randomly sampling 1,000 images rather than using a dedicated validation set.
Nevertheless, we conduct our main experiments using $N=10$, selected based on training set results and achieve the highest fooling rate on the validation set compared to other methods.

\section{Limitations and Discussions}

Applying input transformations in our data-free UAP framework occasionally leads to a decrease in white-box attack performance.
Unlike data-dependent approaches that rely on cross-entropy or logits, our method in Eq.~(8) utilizes activations from all layers.
While this comprehensive use of layer activations provides several advantages, it also increases sensitivity to unintended side effects of input transformations, as shallower features are generally more affected than deeper ones. 
Consequently, although input transformations boost black-box attack transferability, they may cause a slight decline in white-box performance.

In addition, since our method does not rely on target images or models, the adversarial examples generated may exhibit artifacts from the UAP itself, particularly when the images contain large plain regions, making them less visually clean compared to image-specific attacks. 
However, this is not a limitation unique to our approach but a common challenge for UAP methods, where a single UAP is used to attack a wide range of images.

Furthermore, while our method achieves strong performance on CNN architectures, it demonstrates limited attack transferability on ViT-based models.
This limitation appears to be inherent to both data-free and data-dependent UAP approaches.
As a direction for future work, we intend to explore black-box UAP strategies specifically tailored to ViT-based architectures.



%% file: table/add_to_trad.tex
\begin{table*}
\footnotesize
  \centering
  \begin{tabular}{ccccccccc}
    \toprule
    Model & Attack & AN & VGG16 & VGG19 & RN152 & GN & Average\\
    \midrule
    \multirow{2}{*}{RN50} & TRM & 46.46\scriptsize{$\pm$0.80} & 73.82\scriptsize{$\pm$0.85} & 72.43\scriptsize{$\pm$0.91} & 52.64\scriptsize{$\pm$1.18} & 58.59\scriptsize{$\pm$1.57} & 60.79\\
    & \textbf{PSP} & \textbf{51.80}\scriptsize{$\pm$0.80} & \textbf{82.02}\scriptsize{$\pm$0.60} & \textbf{82.09}\scriptsize{$\pm$0.65} & \textbf{60.90}\scriptsize{$\pm$1.14} & \textbf{62.22}\scriptsize{$\pm$1.22} & \textbf{67.81}\\
    \hline

    \multirow{2}{*}{DN121} & TRM & 45.79\scriptsize{$\pm$1.64} & 49.95\scriptsize{$\pm$1.54} & 49.60\scriptsize{$\pm$0.98} & 31.36\scriptsize{$\pm$0.84} & 47.87\scriptsize{$\pm$2.36} & 44.91\\
    & \textbf{PSP} & \textbf{59.04}\scriptsize{$\pm$0.76} & \textbf{67.79}\scriptsize{$\pm$0.85} & \textbf{69.86}\scriptsize{$\pm$0.99} & \textbf{43.72}\scriptsize{$\pm$0.26} & \textbf{72.82}\scriptsize{$\pm$2.04} & \textbf{62.65}\\
    \hline
    
    \multirow{2}{*}{MN-v3} & TRM & 45.47\scriptsize{$\pm$0.49} & 49.13\scriptsize{$\pm$0.71} & 48.69\scriptsize{$\pm$0.64} & 28.67\scriptsize{$\pm$0.58} & 36.15\scriptsize{$\pm$0.92} & 41.62\\
    & \textbf{PSP} &\textbf{66.50}\scriptsize{$\pm$1.30} & \textbf{77.52}\scriptsize{$\pm$0.51} & \textbf{75.96}\scriptsize{$\pm$0.50} & \textbf{49.56}\scriptsize{$\pm$0.72} & \textbf{69.78}\scriptsize{$\pm$0.35} & \textbf{67.86}\\
    \hline
    
    \multirow{2}{*}{Inc-v3} & TRM & \textbf{58.72}\scriptsize{$\pm$0.56} & 71.77\scriptsize{$\pm$0.25} & 70.82\scriptsize{$\pm$0.12} & 45.84\scriptsize{$\pm$0.47} & 62.87\scriptsize{$\pm$0.41} & 62.01\\
    & \textbf{PSP} & 54.84\scriptsize{$\pm$0.55} & \textbf{78.38}\scriptsize{$\pm$0.64} & \textbf{75.52}\scriptsize{$\pm$0.55} & \textbf{52.82}\scriptsize{$\pm$0.54} & \textbf{65.24}\scriptsize{$\pm$0.78} & \textbf{65.36}\\
    \bottomrule
  \end{tabular}
  \caption{Black-box attack transferability across models is analyzed. UAPs crafted on ResNet50, DenseNet121, MobileNet-v3-Large, and Inception-v3 are evaluated on AlexNet, VGG16, VGG19, ResNet152, and GoogleNet.}
  \label{tab:add_to_trad}
\end{table*}

%% file: table/supple_epsilon.tex
\begin{table*}
\footnotesize
  \centering
 \begin{tabular}{cc >{\centering\arraybackslash}p{1.5cm} >{\centering\arraybackslash}p{1.2cm}>{\centering\arraybackslash}p{1.2cm}>{\centering\arraybackslash}p{1.2cm}>{\centering\arraybackslash}p{1.2cm}>{\centering\arraybackslash}p{1.2cm}>{\centering\arraybackslash}p{1.2cm}}
    \toprule
    Model & $\delta_{\infty}$ constraint & Attack~~ & RN50~~~ & DN121~~ & MN-v3-L & Inc-v3~~ & Average~\\
    \midrule
    \multirow{6}{*}{RN50}
    & \multirow{2}{*}{$\epsilon=8$} & TRM-UAP~~~ & 55.39* & 39.80~~~ & 39.02~~~ & 22.87~~~ & 39.27~~~\\
    &  & \textbf{PSP-UAP}~~~ & \textbf{66.41}* & \textbf{50.90}~~~ & \textbf{54.06}~~~ & \textbf{28.98}~~~ & \textbf{50.09}~~~\\
    \cdashline{2-8}[1pt/1pt]
    & \multirow{2}{*}{$\epsilon=10$} & TRM-UAP~~~ & 73.26* & 54.42~~~ & 61.25~~~ & 37.36~~~ & 56.57~~~\\
    &  & \textbf{PSP-UAP}~~~ & \textbf{77.60}* & \textbf{66.11}~~~ & \textbf{70.50}~~~ & \textbf{42.32}~~~ & \textbf{64.13}~~~\\
    \cdashline{2-8}[1pt/1pt]
    & \multirow{2}{*}{$\epsilon=16$}
    & TRM-UAP~~~ & 94.61* & 80.74~~~ & 75.21~~~ & 58.16~~~ & 77.18~~~\\
    &  & \textbf{PSP-UAP}~~~ & \textbf{94.88}* & \textbf{90.53}~~~ & \textbf{90.35}~~~ & \textbf{74.21}~~~ & \textbf{87.49}~~~\\
    \hline

    \multirow{6}{*}{DN121} 
    & \multirow{2}{*}{$\epsilon=8$} & TRM-UAP~~~ & 29.82~~~ & 59.12* & 30.43~~~ & 24.70~~~ & 36.01~~~\\
    &  & \textbf{PSP-UAP}~~~& \textbf{37.56}~~~ & \textbf{67.51}* & \textbf{44.38}~~~ & \textbf{32.34}~~~ & \textbf{45.45}~~~\\
    \cdashline{2-8}[1pt/1pt]
    & \multirow{2}{*}{$\epsilon=10$} & TRM-UAP~~~ & 35.24~~~ & 70.10* & 34.17~~~ & 32.11~~~ & 42.91~~~\\
    &  & \textbf{PSP-UAP}~~~& \textbf{53.03}~~~ & \textbf{85.81}* & \textbf{50.22}~~~ & \textbf{50.73}~~~ & \textbf{59.95}~~~\\
    \cdashline{2-8}[1pt/1pt]
    & \multirow{2}{*}{$\epsilon=16$} 
    & TRM-UAP~~~ & 64.64~~~ & 88.80* & 60.90~~~ & 51.88~~~ & 66.55~~~\\
    &  & \textbf{PSP-UAP}~~~ & \textbf{77.89}~~~ & \textbf{96.84}* & \textbf{77.10}~~~ & \textbf{73.87}~~~ & \textbf{81.42}~~~\\
    \hline

    \multirow{6}{*}{MN-v3-L} 
    & \multirow{2}{*}{$\epsilon=8$} & TRM-UAP~~~ & 37.41~~~ & 36.35~~~ & 79.71* & 30.79~~~ & 46.06~~~ \\
    &  & \textbf{PSP-UAP}~~~ & \textbf{43.47}~~~ & \textbf{44.41}~~~ & \textbf{79.94}* & \textbf{35.39}~~~ & \textbf{50.80}~~~ \\
    \cdashline{2-8}[1pt/1pt]
    & \multirow{2}{*}{$\epsilon=10$} & TRM-UAP~~~ & 39.47~~~ & 40.37~~~ & 73.07* & 30.11~~~ & 45.76~~~ \\
    &  & \textbf{PSP-UAP}~~~ & \textbf{54.38}~~~ & \textbf{54.62}~~~ & \textbf{90.39}* & \textbf{46.29}~~~ & \textbf{61.42}~~~ \\
    \cdashline{2-8}[1pt/1pt]
    & \multirow{2}{*}{$\epsilon=16$}
    & TRM-UAP~~~ & 63.21~~~ & 63.95~~~ & 96.70* & 47.49~~~ & 67.83~~~\\
    &  & \textbf{PSP-UAP}~~~ & \textbf{81.40}~~~ & \textbf{83.45}~~~ & \textbf{99.03}* & \textbf{76.83}~~~ & \textbf{85.18}~~~\\
    \hline

    \multirow{6}{*}{Inc-v3} 
     & \multirow{2}{*}{$\epsilon=8$} & TRM-UAP~~~ & 43.02~~~ & 44.55~~~ & 54.33~~~ & 48.85* & 47.68~~~\\
    &  & \textbf{PSP-UAP}~~~ & \textbf{46.53}~~~ & \textbf{45.43}~~~ & \textbf{57.12}~~~ & \textbf{52.58}* &\textbf{50.41}~~~\\
    \cdashline{2-8}[1pt/1pt]
    & \multirow{2}{*}{$\epsilon=10$} & TRM-UAP~~~ & 53.53~~~ & 54.93~~~ & 67.16~~~ & 64.22* & 59.96~~~\\
    &  & \textbf{PSP-UAP}~~~ & \textbf{57.60}~~~ & \textbf{57.50}~~~ & \textbf{70.20}~~~ & \textbf{65.38}* & \textbf{62.67}~~~\\
    \cdashline{2-8}[1pt/1pt]
    & \multirow{2}{*}{$\epsilon=16$}
    & TRM-UAP~~~ & 78.90~~~ & 79.06~~~ & 88.40~~~ & 91.81* & 84.54~~~\\
    &  & \textbf{PSP-UAP}~~~ & \textbf{83.58}~~~ & \textbf{82.21}~~~& \textbf{89.24}~~~ & \textbf{93.56}* & \textbf{87.14}~~~\\
    \bottomrule
  \end{tabular}
  \caption{$FR$ (\%) results for the UAPs constrained by $\epsilon = 8, 10$ and $16$, crafted on ResNet50, DenseNet121, MobileNet-v3-Large, and Inception-v3. * denotes the white-box model.}
  \label{tab:black_box_epsilonj}
\end{table*}

%% file: table/jpeg.tex

\begin{table}[t]
  \centering
  \footnotesize 
  \setlength{\tabcolsep}{0.1pt} 
  \renewcommand{\arraystretch}{1} 
  \begin{tabular}{l|ccccc|cc}
    \hline
    \multirow{2}{*}{Attack} & \multicolumn{5}{c|}{JPEG compression} & \multicolumn{2}{c}{Ensemble methods} \\
    \cline{2-8}
    & {~AN} & {~~VGG16} & {~~VGG19} & {~RN152} & {~GN}~~ & {Inc-v3$_{ens3}$} & {IncRes-v2$_{ens}$}\\
    \hline
    TRM & ~53.57 & ~~58.38 & ~53.86 & ~39.94* & ~45.44~~ & 17.0 & 11.5 \\
    Ours & ~\textbf{56.74} & ~~\textbf{73.62} &~\textbf{69.41} & ~\textbf{58.61}* & ~\textbf{62.00~~}  & \textbf{19.8} & \textbf{13.1} \\
    \hline
  \end{tabular}
  \caption{Robustness evaluation of our method and TRM against defense methods:  JPEG compression and {ensemble adversarially trained models}.}
  \label{tab:jpeg}
\end{table}

%% file: table/other_models.tex
\begin{table}[t]
  \centering
  \footnotesize 
  \setlength{\tabcolsep}{2.5pt} 
  \renewcommand{\arraystretch}{1} 
  \begin{tabular}{l|cccc|cccc}
    \hline
    Attack & Model & CN-B~~ & DeiT3 & Others & Model & CN-B & DeiT3~~ & Others\\
    \hline
    TRM & \multirow{2}{*}{CN-B}
    & 40.71$^*$ & 10.10 & 34.67 & \multirow{2}{*}{DeiT3} & 14.18 & 6.73$^*$ & 36.01\\
    Ours & & \textbf{86.57}$^*$ & \textbf{13.94} & \textbf{59.49} &  & \textbf{19.98} & \textbf{9.54}$^*$ & \textbf{43.05} \\
    \hline
  \end{tabular}
  \caption{{$FR$ (\%) comparison for ConvNext-B (CN) and DeiT3. \textit{Others} denotes the average $FR$ (\%) on AlexNet, VGG16, VGG19, ResNet152, and GoogleNet.}}
  \vspace{-0.3cm}
  \label{tab:convnext_vit}
\end{table}

%% file: table/temper_tradcnn.tex
\begin{table}
\footnotesize
  \centering
  \footnotesize 
  \setlength{\tabcolsep}{2.5pt} 
  \begin{tabular}{ccccccccc}
    \toprule
    Model & Attack & AN~~ & VGG16~ & VGG19~ & RN152~ & GN~~~ & Avg.\\
    \midrule
    \multirow{3}{*}{AN} & TRM & \textbf{93.53}* & 60.10~~ & 57.08~~ & 27.31~~ & 32.70~~~ & 54.14\\
    & \textbf{PSP-I} & 91.59* & 74.95~~ & 72.70~~ & 47.66~~ & 65.54~~~ & 70.49\\
    & \textbf{PSP-D} & 91.77* & \textbf{76.56}~~ & \textbf{74.07}~~ & \textbf{49.20}~~ & \textbf{66.00}~~ & \textbf{71.52}\\
    \hline

    \multirow{3}{*}{VGG16} & TRM & 47.53~~ & 94.30* & 89.68~~ & 61.43~~ & 53.95~~~ & 69.38\\
    & \textbf{PSP-I} & 48.90~~ & 96.10* & 91.86~~ & 70.75~~ & 58.45~~ & 73.21\\
    & \textbf{PSP-D} & \textbf{50.40}~~ & \textbf{96.26}* & \textbf{92.60}~~ & \textbf{74.10}~~ & \textbf{64.89}~~ & \textbf{75.65} \\
    \hline
    
    \multirow{3}{*}{VGG19} & TRM & 46.01~~ & 89.82~~ & 91.35* & 47.19~~ & 46.48~~ & 64.17\\
    & \textbf{PSP-I} & 46.57~~ & 94.07~~ & 93.88* & 66.08~~ & 57.33~~ & 71.59\\
    & \textbf{PSP-D} & \textbf{48.93}~~ & \textbf{94.55}~~ & \textbf{94.56}* & \textbf{67.13}~~ & \textbf{58.83}~~ & \textbf{72.80} \\
    \hline
    
    \multirow{3}{*}{RN152} & TRM & 53.56~~ & 77.20~~ & 73.30~~ & 67.46* & 57.54~~ & 65.81 \\
    & \textbf{PSP-I} & 57.17~~ & 87.40~~ & 86.34~~ & 84.85* & 71.86~~~ & 77.24\\
    & \textbf{PSP-D} & \textbf{58.82}~~ & \textbf{88.59}~~ & \textbf{87.35}~~ & \textbf{85.65}* & \textbf{76.00}~~ & \textbf{79.29}\\
    \hline
    
    \multirow{3}{*}{GN} & TRM & 60.10~~ & \textbf{79.66}~~ & \textbf{79.98}~~ & \textbf{58.85}~~ & \textbf{85.32}* & \textbf{72.78}\\
    & \textbf{PSP-I} & \textbf{66.06}~~ & 78.88~~ & 79.61~~ & 56.95~~ & 81.04* & 72.51\\
    & \textbf{PSP-D} & 65.22~~ & 78.43~~ & 79.26~~ & 57.63~~ & 81.43* & 72.39\\

    \bottomrule
  \end{tabular}
  \caption{Ablation study on the sample reweighting temperature parameter, $\tau$.
  PSP-I and PSP-D refer to fixing the $\tau$ to 4 and adapting it for each model.}
  \label{tab:temper_tradcnn}
\end{table}

%% file: table/temper_addcnn.tex
\begin{table}
\footnotesize
  \centering
  \begin{tabular}{ccccccc}
    \toprule
    Model & Attack & RN50~~ & DN121~~ & MN-v3~ & Inc-v3~~ & Avg.\\
    \midrule
    \multirow{3}{*}{RN50} & TRM & 73.26* & 54.42~~ & 61.25~~ & 37.36~~ & 56.57\\
    & \textbf{PSP-I} & 76.41* & 64.89~~ & 69.32~~ & 42.03~~ & 63.16\\
    & \textbf{PSP-D} & \textbf{77.60}* & \textbf{66.11}~~ & \textbf{70.50}~~ & \textbf{42.32}~~ & \textbf{64.13}\\
    \hline
    
    \multirow{3}{*}{DN121} & TRM & 35.24~~ & 70.10* & 34.17~~ & 32.11~~ & 42.91\\
    & \textbf{PSP-I} & \textbf{53.30}~~ & 84.95* & 49.79~~ & 49.59~~ & 59.40\\
    & \textbf{PSP-D} & 53.03~~ & \textbf{85.81}* & \textbf{50.22}~~ & \textbf{50.73}~~ & \textbf{59.95}\\
    \hline
    
    \multirow{3}{*}{MN-v3} & TRM & 39.47~~ & 40.37~~ & 73.07* & 30.11~~ & 45.76\\
    & \textbf{PSP-I} & \textbf{54.88}~~ & 53.56~~ & 89.85* & 45.92~~ & 61.05\\
    & \textbf{PSP-D} & 54.38~~ & \textbf{54.62}~~ & \textbf{90.39}* & \textbf{46.29}~~ & \textbf{61.42}\\
    \hline
    
    \multirow{3}{*}{Inc-v3} & TRM & 53.53~~ & 54.93~~ & 67.16~~ & 64.22* & 59.96\\
    & \textbf{PSP-I} & 57.56~~ & 57.15~~ & 69.94~~ & 64.83* & 62.37\\
    & \textbf{PSP-D} & \textbf{57.60}~~ & \textbf{57.50}~~ & \textbf{70.20}~~ & \textbf{65.38}* & \textbf{62.67}\\

    \bottomrule
  \end{tabular}
  \caption{Ablation study on the sample reweighting temperature parameter, $\tau$, for additional CNN models. PSP-I and PSP-D refer to fixing the $\tau$ to 4 and adapting it for each model.}
  \label{tab:black_box_temp_2}
\end{table}

%% file: main.bbl
\begin{thebibliography}{47}
\providecommand{\natexlab}[1]{#1}
\providecommand{\url}[1]{\texttt{#1}}
\expandafter\ifx\csname urlstyle\endcsname\relax
  \providecommand{\doi}[1]{doi: #1}\else
  \providecommand{\doi}{doi: \begingroup \urlstyle{rm}\Url}\fi

\bibitem[Bertinetto et~al.(2016)Bertinetto, Valmadre, Henriques, Vedaldi, and Torr]{siam}
Luca Bertinetto, Jack Valmadre, Joao~F Henriques, Andrea Vedaldi, and Philip~HS Torr.
\newblock Fully-convolutional siamese networks for object tracking.
\newblock In \emph{ECCV Workshops}, 2016.

\bibitem[Carlini and Wagner(2017)]{security}
Nicholas Carlini and David Wagner.
\newblock Towards evaluating the robustness of neural networks.
\newblock In \emph{IEEE Symposium on Security and Privacy (SP)}, 2017.

\bibitem[Dong et~al.(2018)Dong, Liao, Pang, Su, Zhu, Hu, and Li]{mifgsm}
Yinpeng Dong, Fangzhou Liao, Tianyu Pang, Hang Su, Jun Zhu, Xiaolin Hu, and Jianguo Li.
\newblock Boosting adversarial attacks with momentum.
\newblock In \emph{CVPR}, 2018.

\bibitem[Dong et~al.(2019)Dong, Pang, Su, and Zhu]{dong2019evading}
Yinpeng Dong, Tianyu Pang, Hang Su, and Jun Zhu.
\newblock Evading defenses to transferable adversarial examples by translation-invariant attacks.
\newblock In \emph{CVPR}, 2019.

\bibitem[Eykholt et~al.(2018)Eykholt, Evtimov, Fernandes, Li, Rahmati, Xiao, Prakash, Kohno, and Song]{autonomous}
Kevin Eykholt, Ivan Evtimov, Earlence Fernandes, Bo Li, Amir Rahmati, Chaowei Xiao, Atul Prakash, Tadayoshi Kohno, and Dawn Song.
\newblock Robust physical-world attacks on deep learning visual classification.
\newblock In \emph{CVPR}, 2018.

\bibitem[Goodfellow et~al.(2015)Goodfellow, Shlens, and Szegedy]{FGSM}
Ian~J Goodfellow, Jonathon Shlens, and Christian Szegedy.
\newblock Explaining and harnessing adversarial examples.
\newblock In \emph{ICLR}, 2015.

\bibitem[Guo et~al.(2018)Guo, Rana, Cisse, and van~der Maaten]{jpeg}
Chuan Guo, Mayank Rana, Moustapha Cisse, and Laurens van~der Maaten.
\newblock Countering adversarial images using input transformations.
\newblock In \emph{ICLR}, 2018.

\bibitem[He et~al.(2016)He, Zhang, Ren, and Sun]{res152}
Kaiming He, Xiangyu Zhang, Shaoqing Ren, and Jian Sun.
\newblock Deep residual learning for image recognition.
\newblock In \emph{CVPR}, 2016.

\bibitem[Howard et~al.(2019)Howard, Sandler, Chu, Chen, Chen, Tan, Wang, Zhu, Pang, Vasudevan, et~al.]{mobile}
Andrew Howard, Mark Sandler, Grace Chu, Liang-Chieh Chen, Bo Chen, Mingxing Tan, Weijun Wang, Yukun Zhu, Ruoming Pang, Vijay Vasudevan, et~al.
\newblock Searching for mobilenetv3.
\newblock In \emph{CVPR}, 2019.

\bibitem[Huang et~al.(2017)Huang, Liu, Van Der~Maaten, and Weinberger]{dense}
Gao Huang, Zhuang Liu, Laurens Van Der~Maaten, and Kilian~Q Weinberger.
\newblock Densely connected convolutional networks.
\newblock In \emph{CVPR}, 2017.

\bibitem[Krizhevsky et~al.(2012)Krizhevsky, Sutskever, and Hinton]{alex}
Alex Krizhevsky, Ilya Sutskever, and Geoffrey~E Hinton.
\newblock Imagenet classification with deep convolutional neural networks.
\newblock In \emph{NeurIPS}, 2012.

\bibitem[Kurakin et~al.(2018)Kurakin, Goodfellow, and Bengio]{i-fgsm}
Alexey Kurakin, Ian~J Goodfellow, and Samy Bengio.
\newblock Adversarial examples in the physical world.
\newblock In \emph{Artificial intelligence safety and security}. Chapman and Hall/CRC, 2018.

\bibitem[Li et~al.(2022)Li, Yang, Wei, Yang, and Huang]{at-uap}
Maosen Li, Yanhua Yang, Kun Wei, Xu Yang, and Heng Huang.
\newblock Learning universal adversarial perturbation by adversarial example.
\newblock In \emph{AAAI}, 2022.

\bibitem[Lin et~al.(2020)Lin, Song, He, Wang, and Hopcroft]{lin2019nesterov}
Jiadong Lin, Chuanbiao Song, Kun He, Liwei Wang, and John~E Hopcroft.
\newblock Nesterov accelerated gradient and scale invariance for adversarial attacks.
\newblock In \emph{ICLR}, 2020.

\bibitem[Liu et~al.(2019)Liu, Ji, Li, Zhang, Gao, Wu, and Huang]{pd-ua}
Hong Liu, Rongrong Ji, Jie Li, Baochang Zhang, Yue Gao, Yongjian Wu, and Feiyue Huang.
\newblock Universal adversarial perturbation via prior driven uncertainty approximation.
\newblock In \emph{ICCV}, 2019.

\bibitem[Liu et~al.(2023{\natexlab{a}})Liu, Zhong, Zhang, Qin, and Deng]{sga-uap}
Xuannan Liu, Yaoyao Zhong, Yuhang Zhang, Lixiong Qin, and Weihong Deng.
\newblock Enhancing generalization of universal adversarial perturbation through gradient aggregation.
\newblock In \emph{ICCV}, 2023{\natexlab{a}}.

\bibitem[Liu et~al.(2023{\natexlab{b}})Liu, Feng, Wang, Yang, and Ming]{TRM}
Yiran Liu, Xin Feng, Yunlong Wang, Wu Yang, and Di Ming.
\newblock Trm-uap: Enhancing the transferability of data-free universal adversarial perturbation via truncated ratio maximization.
\newblock In \emph{ICCV}, 2023{\natexlab{b}}.

\bibitem[Liu et~al.(2022)Liu, Mao, Wu, Feichtenhofer, Darrell, and Xie]{convnext}
Zhuang Liu, Hanzi Mao, Chao-Yuan Wu, Christoph Feichtenhofer, Trevor Darrell, and Saining Xie.
\newblock A convnet for the 2020s.
\newblock In \emph{CVPR}, 2022.

\bibitem[Madry et~al.(2018)Madry, Makelov, Schmidt, Tsipras, and Vladu]{pgd}
Aleksander Madry, Aleksandar Makelov, Ludwig Schmidt, Dimitris Tsipras, and Adrian Vladu.
\newblock Towards deep learning models resistant to adversarial attacks.
\newblock In \emph{ICLR}, 2018.

\bibitem[Moosavi-Dezfooli et~al.(2016)Moosavi-Dezfooli, Fawzi, and Frossard]{Deepfool}
Seyed-Mohsen Moosavi-Dezfooli, Alhussein Fawzi, and Pascal Frossard.
\newblock Deepfool: a simple and accurate method to fool deep neural networks.
\newblock In \emph{CVPR}, 2016.

\bibitem[Moosavi-Dezfooli et~al.(2017)Moosavi-Dezfooli, Fawzi, Fawzi, and Frossard]{UAP}
Seyed-Mohsen Moosavi-Dezfooli, Alhussein Fawzi, Omar Fawzi, and Pascal Frossard.
\newblock Universal adversarial perturbations.
\newblock In \emph{CVPR}, 2017.

\bibitem[Mopuri et~al.(2017)Mopuri, Garg, and Venkatesh~Babu]{FFF}
KR Mopuri, U Garg, and R Venkatesh~Babu.
\newblock Fast feature fool: A data independent approach to universal adversarial perturbations.
\newblock In \emph{BMVC}, 2017.

\bibitem[Mopuri et~al.(2018{\natexlab{a}})Mopuri, Ganeshan, and Babu]{gd-uap}
Konda~Reddy Mopuri, Aditya Ganeshan, and R~Venkatesh Babu.
\newblock Generalizable data-free objective for crafting universal adversarial perturbations.
\newblock \emph{TPAMI}, 41\penalty0 (10), 2018{\natexlab{a}}.

\bibitem[Mopuri et~al.(2018{\natexlab{b}})Mopuri, Ojha, Garg, and Babu]{nag}
Konda~Reddy Mopuri, Utkarsh Ojha, Utsav Garg, and R~Venkatesh Babu.
\newblock Nag: Network for adversary generation.
\newblock In \emph{CVPR}, 2018{\natexlab{b}}.

\bibitem[Mopuri et~al.(2018{\natexlab{c}})Mopuri, Uppala, and Babu]{AAA}
Konda~Reddy Mopuri, Phani~Krishna Uppala, and R~Venkatesh Babu.
\newblock Ask, acquire, and attack: Data-free uap generation using class impressions.
\newblock In \emph{ECCV}, 2018{\natexlab{c}}.

\bibitem[Poursaeed et~al.(2018)Poursaeed, Katsman, Gao, and Belongie]{gap}
Omid Poursaeed, Isay Katsman, Bicheng Gao, and Serge Belongie.
\newblock Generative adversarial perturbations.
\newblock In \emph{CVPR}, 2018.

\bibitem[Redmon(2016)]{yolo}
J Redmon.
\newblock You only look once: Unified, real-time object detection.
\newblock In \emph{CVPR}, 2016.

\bibitem[Ren et~al.(2016)Ren, He, Girshick, and Sun]{faster}
Shaoqing Ren, Kaiming He, Ross Girshick, and Jian Sun.
\newblock Faster r-cnn: Towards real-time object detection with region proposal networks.
\newblock \emph{TPAMI}, 39\penalty0 (6):\penalty0 1137--1149, 2016.

\bibitem[Ronneberger et~al.(2015)Ronneberger, Fischer, and Brox]{u-net}
Olaf Ronneberger, Philipp Fischer, and Thomas Brox.
\newblock U-net: Convolutional networks for biomedical image segmentation.
\newblock In \emph{MICCAI}, 2015.

\bibitem[Russakovsky et~al.(2015)Russakovsky, Deng, Su, Krause, Satheesh, Ma, Huang, Karpathy, Khosla, Bernstein, et~al.]{Imagenet}
Olga Russakovsky, Jia Deng, Hao Su, Jonathan Krause, Sanjeev Satheesh, Sean Ma, Zhiheng Huang, Andrej Karpathy, Aditya Khosla, Michael Bernstein, et~al.
\newblock Imagenet large scale visual recognition challenge.
\newblock \emph{IJCV}, 115, 2015.

\bibitem[Selvaraju et~al.(2017)Selvaraju, Cogswell, Das, Vedantam, Parikh, and Batra]{gradcam}
Ramprasaath~R Selvaraju, Michael Cogswell, Abhishek Das, Ramakrishna Vedantam, Devi Parikh, and Dhruv Batra.
\newblock Grad-cam: Visual explanations from deep networks via gradient-based localization.
\newblock In \emph{ICCV}, 2017.

\bibitem[Shafahi et~al.(2020)Shafahi, Najibi, Xu, Dickerson, Davis, and Goldstein]{spgd}
Ali Shafahi, Mahyar Najibi, Zheng Xu, John Dickerson, Larry~S Davis, and Tom Goldstein.
\newblock Universal adversarial training.
\newblock In \emph{AAAI}, 2020.

\bibitem[Simonyan and Zisserman(2015)]{vgg}
Karen Simonyan and Andrew Zisserman.
\newblock Very deep convolutional networks for large-scale image recognition.
\newblock In \emph{ICLR}, 2015.

\bibitem[Szegedy et~al.(2014)Szegedy, Zaremba, Sutskever, Bruna, Erhan, Goodfellow, and Fergus]{l-bfgs}
Christian Szegedy, Wojciech Zaremba, Ilya Sutskever, Joan Bruna, Dumitru Erhan, Ian Goodfellow, and Rob Fergus.
\newblock Intriguing properties of neural networks.
\newblock In \emph{ICLR}, 2014.

\bibitem[Szegedy et~al.(2015)Szegedy, Liu, Jia, Sermanet, Reed, Anguelov, Erhan, Vanhoucke, and Rabinovich]{google}
Christian Szegedy, Wei Liu, Yangqing Jia, Pierre Sermanet, Scott Reed, Dragomir Anguelov, Dumitru Erhan, Vincent Vanhoucke, and Andrew Rabinovich.
\newblock Going deeper with convolutions.
\newblock In \emph{CVPR}, 2015.

\bibitem[Szegedy et~al.(2016{\natexlab{a}})Szegedy, Vanhoucke, Ioffe, Shlens, and Wojna]{inc_v3}
Christian Szegedy, Vincent Vanhoucke, Sergey Ioffe, Jon Shlens, and Zbigniew Wojna.
\newblock Rethinking the inception architecture for computer vision.
\newblock In \emph{CVPR}, 2016{\natexlab{a}}.

\bibitem[Szegedy et~al.(2016{\natexlab{b}})Szegedy, Vanhoucke, Ioffe, Shlens, and Wojna]{inception}
Christian Szegedy, Vincent Vanhoucke, Sergey Ioffe, Jon Shlens, and Zbigniew Wojna.
\newblock Rethinking the inception architecture for computer vision.
\newblock In \emph{CVPR}, 2016{\natexlab{b}}.

\bibitem[Touvron et~al.(2022)Touvron, Cord, and J{\'e}gou]{deit}
Hugo Touvron, Matthieu Cord, and Herv{\'e} J{\'e}gou.
\newblock Deit iii: Revenge of the vit.
\newblock In \emph{ECCV}. Springer, 2022.

\bibitem[Tram{\`e}r et~al.(2018)Tram{\`e}r, Kurakin, Papernot, Goodfellow, Boneh, and McDaniel]{ens_adv}
Florian Tram{\`e}r, Alexey Kurakin, Nicolas Papernot, Ian Goodfellow, Dan Boneh, and Patrick McDaniel.
\newblock Ensemble adversarial training: Attacks and defenses.
\newblock In \emph{ICLR}, 2018.

\bibitem[Wang et~al.(2024)Wang, He, Wang, and Wang]{bsr}
Kunyu Wang, Xuanran He, Wenxuan Wang, and Xiaosen Wang.
\newblock Boosting adversarial transferability by block shuffle and rotation.
\newblock In \emph{CVPR}, 2024.

\bibitem[Wang et~al.(2021)Wang, He, Wang, and He]{admix}
Xiaosen Wang, Xuanran He, Jingdong Wang, and Kun He.
\newblock Admix: Enhancing the transferability of adversarial attacks.
\newblock In \emph{ICCV}, 2021.

\bibitem[Wang et~al.(2023)Wang, Zhang, and Zhang]{SIA}
Xiaosen Wang, Zeliang Zhang, and Jianping Zhang.
\newblock Structure invariant transformation for better adversarial transferability.
\newblock In \emph{ICCV}, 2023.

\bibitem[Xiao et~al.(2018)Xiao, Li, Zhu, He, Liu, and Song]{generating}
Chaowei Xiao, Bo Li, Jun-Yan Zhu, Warren He, Mingyan Liu, and Dawn Song.
\newblock Generating adversarial examples with adversarial networks.
\newblock In \emph{IJCAI}, 2018.

\bibitem[Xie et~al.(2019)Xie, Zhang, Zhou, Bai, Wang, Ren, and Yuille]{xie2019improving}
Cihang Xie, Zhishuai Zhang, Yuyin Zhou, Song Bai, Jianyu Wang, Zhou Ren, and Alan~L Yuille.
\newblock Improving transferability of adversarial examples with input diversity.
\newblock In \emph{CVPR}, 2019.

\bibitem[Zhang et~al.(2021{\natexlab{a}})Zhang, Benz, Karjauv, and Kweon]{cosine}
Chaoning Zhang, Philipp Benz, Adil Karjauv, and In~So Kweon.
\newblock Data-free universal adversarial perturbation and black-box attack.
\newblock In \emph{ICCV}, 2021{\natexlab{a}}.

\bibitem[Zhang et~al.(2021{\natexlab{b}})Zhang, Wang, Wang, Zeng, and Liu]{fairmot}
Yifu Zhang, Chunyu Wang, Xinggang Wang, Wenjun Zeng, and Wenyu Liu.
\newblock Fairmot: On the fairness of detection and re-identification in multiple object tracking.
\newblock \emph{IJCV}, 129:\penalty0 3069--3087, 2021{\natexlab{b}}.

\bibitem[Zhu et~al.(2024)Zhu, Zhang, Liang, Liu, and Xu]{L2T}
Rongyi Zhu, Zeliang Zhang, Susan Liang, Zhuo Liu, and Chenliang Xu.
\newblock Learning to transform dynamically for better adversarial transferability.
\newblock In \emph{CVPR}, 2024.

\end{thebibliography}
